\documentclass{article}

\usepackage[english]{babel}

\usepackage[letterpaper,top=2cm,bottom=2cm,left=3cm,right=3cm,marginparwidth=1.75cm]{geometry}

\usepackage{amsmath}
\usepackage{graphicx}
\usepackage[colorlinks=true, allcolors=blue]{hyperref}
\usepackage{enumitem}
\usepackage{soul}

\usepackage{color} 
\title{Toward AI-Driven Digital Organism: \\ A System of Multiscale Foundation Models for Predicting, Simulating and Programming Biology at All Levels}
\author{Le Song$^{\star,\diamond}$, Eran Segal$^{\star,\diamond,\dagger}$, Eric Xing$^{\star,\diamond,\ddagger}$ 
\\
\\
$^\star$GenBio AI \\
\\
$^\diamond$Mohamed bin Zayed University of Artificial Intelligence \\
$^\dagger$Weizmann Institute of Science  \\
$^\ddagger$Carnegie Mellon University\\ \\
\{le.song, eran.segal, eric.xing\}@genbio.ai 
}

\begin{document}

\date{}

\maketitle

\begin{abstract}

We present an approach of using AI to model and simulate biology and life. Why is it important? Because at the core of medicine, pharmacy, public health, longevity, agriculture and food security, environmental protection, and clean energy, it is biology at work. Biology in the physical world is too complex to manipulate and always expensive and risky to tamper with. In this perspective, we layout an engineering viable approach to address this challenge by constructing an AI-Driven Digital Organism (AIDO), a system of integrated multiscale foundation models, in a modular, connectable, and holistic fashion to reflect biological scales, connectedness, and complexities. An AIDO opens up a safe, affordable and high-throughput alternative platform for predicting, simulating and programming biology at all levels from molecules to cells to individuals. We envision that an AIDO is poised to trigger a new wave of better-guided wet-lab experimentation and better-informed first-principle reasoning, which can eventually help us better decode and improve life.  

\end{abstract}


\section{Introduction} Biology lies at the core of vital fields such as medicine, pharmacy, public health, longevity, agriculture and food security, environmental protection, and clean energy. The mechanisms underlying living and physical systems have always fascinated us. With Newton’s laws, we can predict the orbits of celestial bodies; the periodic table allows us to anticipate the properties of chemical compounds; and we can even simulate weather and environmental systems. However, despite our extensive knowledge of atomic, molecular, chemical, and physical laws, and the computational power of modern computers, we still cannot simulate biological systems accurately. Whether we aim to pinpoint genetic markers of diseases for diagnosis, design drugs to heal damaged cells or deter pathogens, or develop vaccines to combat pandemics, such advancements in medicine consistently require a profound understanding of the underlying biology at all levels, along with the ability to predict, simulate, and program biological activities comprehensively. Manipulating biology in the physical world is extremely complex, expensive, and risky, and should be preceded by extensive computer-aided digital design, simulation, and validation as in other industrial fields such as civil, nuclear, and semiconductor engineering. We propose a vision in which such capabilities can be realized using generative AI.

Generative AI and large pretrained models across text, images, speech, and video have become key pillars for advancing artificial general intelligence (AGI), driving significant improvements in a wide range of downstream tasks, including language and image comprehension, translation, knowledge extraction, reasoning, and cross-modal generation. 
These models are often known as ``foundation models" (FMs)~\cite{bommasani2021opportunities}, because of their relatively standardized architecture across data and utilities, generalizability over different applications, and their massive scale and cost of production. 
While these models excel as generalists—handling everyday human communication tasks such as casual conversations, answering questions about locations, celebrities, history, travel, and other basic information, and advancing into more challenging tasks such as logical reasoning and planning—they fall short in domains such as biology where the laws of nature at all scales and dimensions beyond human experiences are at work, particularly in understanding complex phenomena like cellular activities and generating scientific knowledge and hypotheses based on exceedingly complex biological data. 

One of the major challenges in building foundation models for biology is that biological and life science problems operate in a language vastly different from natural languages and images. They encompass multiscale complexities spanning from the molecular level (DNA, RNA, and proteins), through network levels (protein interaction networks, regulatory networks, and gene expression within cells), to intricate systems like cell-cell interactions, organs, individuals, and societies. Historically, numerous specialized machine learning and computational biology models have been developed to address specific issues within various facets of biology and life sciences. However, these models are often constructed with limited labeled data and tailored for particular applications—well-known as the so-called ``one-model for one-task" principle, resulting in uncontextualized and often suboptimal performance and limited transferability to other problems.

\begin{figure}
    \centering
    \includegraphics[width=\textwidth]{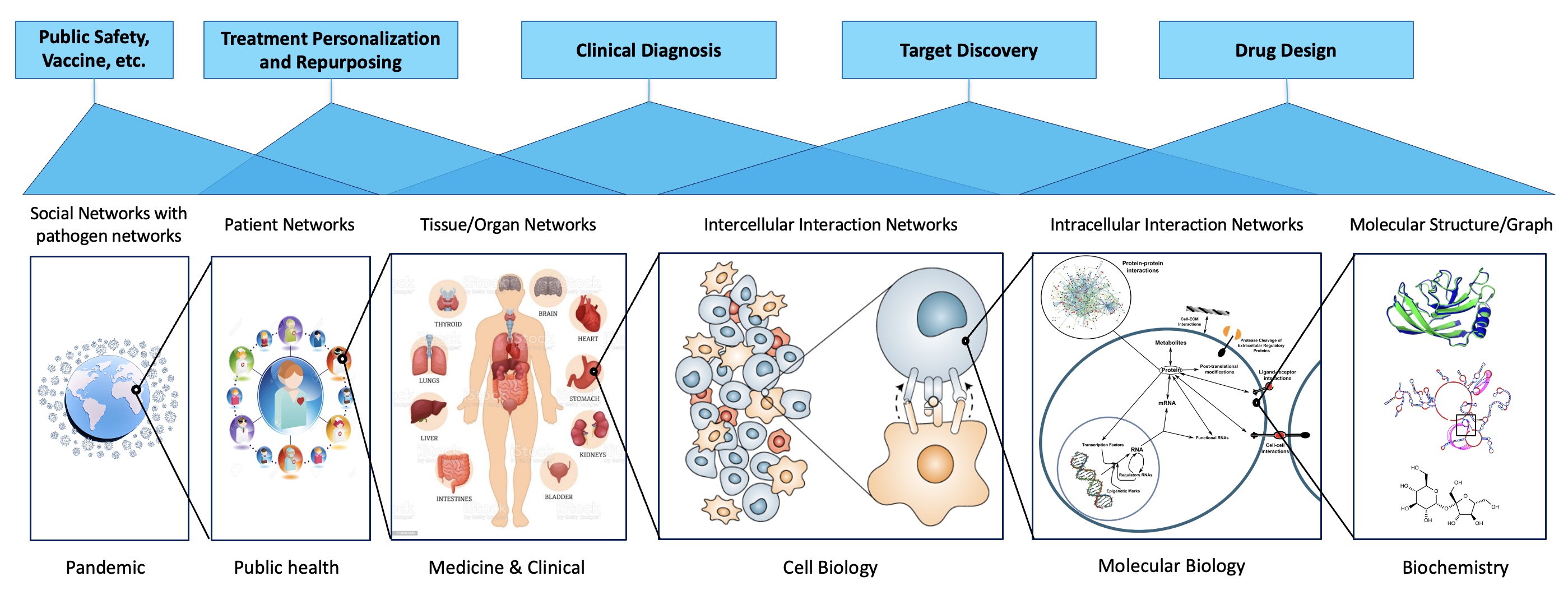}
    \caption{Biology is a complex multiscale network.}
    \label{fig:multiscalebio}
\end{figure}

In recent years, there has been a notable surge in efforts within academia and industry to explore the opportunities presented by pretrained large models, or foundation models, for biology and life sciences. For DNA sequences, models like Nucleotide Transformer~\cite{dalla2023nucleotide} and HyenaDNA~\cite{nguyen2024hyenadna} have been developed, leading to improvements in a range of genome-related downstream tasks. Similarly, RNA-FM~\cite{chen2022interpretable} and CodonBERT~\cite{li2023codonbert} have been created for RNA analysis. In the realm of protein sequences, models such as ESM~\cite{lin2023evolutionary}, ProGen2~\cite{nijkamp2023progen2}, and xTrimoPGLM~\cite{chen2024xtrimopglm} have been introduced. Advances in protein structure prediction have been achieved with models like AlphaFold~\cite{jumper2021highly,abramson2024accurate}, ESMFold~\cite{lin2023evolutionary}, and GearNet~\cite{zhang2022protein}. For single-cell RNA sequencing, models such as GeneFormer~\cite{theodoris2023transfer}, scFoundation~\cite{hao2024large}, and scGPT~\cite{cui2024scgpt} have been developed. Additionally, BioGPT~\cite{luo2022biogpt} and MedSAM~\cite{ma2024segment} have been applied to biomedical text documents and images. The amount of data available for pretraining these models has grown beyond billions of data points, and model sizes have risen to internet-scale levels with parameters exceeding 100 billion~\cite{chen2024xtrimopglm}.

Despite recent advancements, most current foundation models in biology are designed and built for individual data modalities. They do not account for the multiscale nature of biology and the multimodal characteristics of biological data, and therefore, are not quite "foundational" for biology. Consequently, a truly {\it foundational} model which is intrinsically integrative, multiscale, and connected across diverse biological data, and is capable of addressing biological questions across different scales is still missing. 
It is our view that, a foundation model for biology—which can be a system of component FMs—needs to incorporate multiple types of data and biological constraints arising from different biological scales. Furthermore, such a system is more than just an agglomeration of modality-specific FMs, and must involve system-wide harmonization through nested or hierarchical representation propagation, utilization, fine-tuning, or continual pretraining. It should also have the ability to connect different FM modules from the system, and provide a foundation to address more complex prediction, simulation, and reprogramming tasks arising from molecules, cells, organisms, and beyond. We coin such an integrated multiscale system of foundation models an \textbf{\textit{AI-Driven Digital Organism (AIDO)}}. 

In constructing the AIDO, several key challenges need to be addressed. First, it is essential to determine what constitutes a valid set of component foundation models within an AIDO, and how to achieve a parsimonious yet gapless collection. This involves deciding which models are necessary to capture the complexity of biological systems and ensuring that they cover a broad spectrum of biological phenomena. Second, constructing these models requires careful consideration of data acquisition and preprocessing. The availability and quality of biological data significantly impact the performance and applicability of the foundation models. Third, selecting suitable deep learning architectures is vital for developing effective foundation models constituting the AIDO. It is necessary to explore which types of architectures are most appropriate for capturing the intricate patterns and relationships inherent in specific biological data modalities. Moreover, integrating biological knowledge into these models poses a significant challenge. Effective approaches must be developed to incorporate domain-specific knowledge, ensuring that the models not only learn from data but also adhere to known biological principles. This integration can enhance the models’ interpretability and reliability. Additionally, incorporating the multiscale and multimodal nature of biology is essential for connecting different pretrained component FMs. Developing methods that integrate data from various scales and modalities will enable the models to capture the interconnectedness of biological systems, from molecular interactions to cellular processes and organismal behaviors. Determining suitable pretraining tasks for these models is another important consideration. The choice of pretraining tasks influences the models’ ability to generalize and perform well on downstream tasks. These tasks should be designed to capture the fundamental aspects of biological data and prepare the models for a wide range of applications. Furthermore, it is necessary to identify which downstream tasks can benefit from these pretrained models and to what extent. This includes evaluating how much improvement can be achieved in tasks such as disease diagnosis, drug discovery, and understanding cellular mechanisms. Addressing these challenges is crucial for developing a robust AIDO that can enhance our ability to predict, simulate, and program biological activities across different scales. 

In this perspective, we aim to lay out a blueprint to address these challenges, offering our insights on how to construct an AIDO in three phases: {\it divide and conquer}, where we build the collection of all component FMs separately on specific data; {\it connect the dots}, where we integrate component FMs across modalities and scales; and finally, {\it piece it all together}, where we align and optimize all components holistically. Rather than focusing on specific biological problems or data modalities as in current FM4Bio research, we propose advancing an integrative system of component FMs spanning multiple scales enabling greater generality and transferability across diverse downstream tasks in a one-stop turnkey manner. 
We believe that by further advancing along this vision, an AIDO can be created and continuously augmented, enabling us to address a broad range of biological, medical, and pharmaceutical problems computationally, and simulate various types of phenomena arising from different scales and aspects within life sciences and the health industry.

\section{Multiscale Structure and Organization of Biological Systems}

Biological systems are organized as multiscale, heterogeneous networks of interacting entities, ranging from molecules to cells to organisms within their environments (see text books such as~\cite{chaffey2003alberts} for more details; and see  Figure~\ref{fig:multiscalebio} for an illustration). At the most fundamental level, these entities are various types of molecules and their interactions. Key molecules such as DNA, RNA, and proteins operate under the central dogma of molecular biology, where DNA encodes genetic information, RNA serves as the intermediary, and proteins execute cellular functions.

DNA sequences comprise coding regions, regulatory elements like enhancers and promoters, and noncoding intergenic regions. Through transcription, DNA is expressed as RNA sequences. While some RNA molecules are noncoding and perform regulatory functions or act directly within the cell, messenger RNA (mRNA) is translated by ribosomes into protein sequences. These proteins fold into specific three-dimensional (3D) structures to carry out diverse functions, including regulation, signaling, enzymatic activity, scaffolding, and transport.

Within the cellular environment, molecules and ions form complex, stochastic interaction networks involving protein-protein, protein-DNA, protein-RNA, protein-small molecule, and protein-ion interactions. The functions and interactions of these molecules are often governed by their structures and physicochemical properties. For example, the 3D organization of DNA within the nucleus influences gene accessibility and expression; protein-protein interactions require complementary shapes and charges; and enzymatic reactions depend on the precise spatial arrangement of amino acid residues relative to substrates.

These molecular interaction networks constitute complex systems exhibiting dynamic behaviors determined by initial states, governing biochemical equations, and external stimuli or perturbations. Dynamic cellular processes such as signal transduction and cell division arise from these intricate networks. Cells interact with one another, especially when in direct contact or close proximity, through mechanisms that include membrane protein interactions, cytokine and hormone signaling, and the exchange of ions or small molecules. Such cell-cell interactions couple the individual dynamic systems of cells, leading to complex spatial patterns of molecular distributions and coordinated tissue functions.

Spatially organized assemblies of functionally diverse cells form tissues and organs. The coordinated arrangement and interaction of cells within these structures are essential for the normal function of tissues and organs. At the organismal level, tissues and organs coordinate to form self-sustaining individuals capable of interacting with their environment. Organisms exchange chemical and biological substances with their surroundings and with each other, forming networks of biological entities. Environmental factors, in turn, influence the physiological conditions and adaptive responses of these organisms. Understanding this multiscale structure and organization is crucial for developing comprehensive models that can simulate biological phenomena across different scales. 

Given the multiscale organization of biological entities, it is essential to model and utilize data across appropriate scales of granularity to effectively address specific biological questions. For example, clinical diagnosis relies on individual-level data spanning molecular to system granularity —including phenotypic measurements along with genetic and cellular information—to enable personalized medicine and accurate disease identification. Understanding disease mechanisms and discovering therapeutic targets necessitate examining tissues, cells, and molecular interactions to reveal underlying pathways and dysfunctions. Drug design focuses on molecular and pathway data to identify compounds that precisely modulate biological processes or protein functions.

In fields like developmental biology, integrating data across scales—from gene expression within cells to the coordinated behavior of tissues and organs—is crucial for deciphering growth processes and correcting abnormalities. Therefore, addressing questions at different levels requires a system of foundation models that are themselves multiscale and multimodal. Such models should represent and integrate data across various biological scales, providing a comprehensive framework for understanding and manipulating biological systems, and comprising an AIDO.

\section{What is an AI-Driven Digital Organism}

In our vision, a digital organism is a computational model—such as a transformer (or newer architectures to emerge)-based foundation model—that enables the simulation of all biological, physiological, and clinical events occurring within a living organism. This digital organism should be consistent with biogical scales, connectedness and complexities, and constructed using multimodal, large-scale datasets, including molecular sequences and structures, biological networks and pathways, transcriptomic and metabolomic profiles, images, textual descriptions, and spatial-temporal information of biological systems. It should model a living system in a multiscale manner, encompassing molecular, genetic, structural, cellular, tissue, organ, organismal, populational, and evolutionary levels.

In the following, we overview more concisely a set of desiderata for what we expect from a digital organism and how it is different from other approaches. 

\subsection{Universal Representations of Biological Entities}

At the core of the digital organism concept is the representation of biological entities at various levels—including genes, operons, regulatory elements, proteins, organelles, cells, tissues, and entire organisms—using explicit, operationalizable, and multi-resolution digital expressions such as vectors or tensors. These representations are derived from encoders that compute latent states at the desired level of granularity from raw input data associated with the biological subject. They can be either pre-computed and deposited to a repository bank, or computed live prompted by new data and unique context in novel situations. This process can also incorporate additional context, including influences from interacting entities, temporal-spatial conditions, and prior knowledge. With such representations, a wide range of downstream predictive, simulative, and programmatic applications can be facilitated as detailed later in this paper. 
These representations can help mitigate issues related to limited labeled data for high dimensional inputs and to transfer knowledge across tasks making them foundational and instrumental in addressing a wide spectrum of downstream problems involving similar types of inputs.

Key to their flexibility and versatility in these applications, these digital representations are amenable to a wide range of computational operations that enhance their utility:

\textbf{Arithmetic Operations:} These operations allow for the combination or subtraction of representations across multiple entities that are at the same level or scale, facilitating comparative studies and differential analyses. For example, adding or subtracting gene expression vectors can highlight upregulated or downregulated pathways between healthy and diseased states. Amplifying or attenuating signals within these vectors can simulate the effects of dosage variations, electrical conductivity changes, or environmental influences on biological processes. Concatenation or truncation operations can model the integration or loss of biological effects, such as in gene fusion events or alternative splicing variants. 

\textbf{Machine Learning Operations:} Applying clustering or dimensional reduction algorithms to these representations can reveal natural groupings within the data, such as revealing cell types based on gene expression profiles. Classification models can assign labels to unknown samples, aiding in tasks like disease diagnosis. Predictive modeling can forecast biological outcomes, such as predicting protein folding structures from amino acid sequences. The representations of the biological entities can be used individually or in combination for specific downstream predictive or generative tasks by fine-tuning and adapting with a small number of labeled data points. This approach often results in significant improvements in accuracy and convergence speed compared to training models from scratch.
Temporal and spatial processing techniques enable the modeling of dynamic biological processes over time and across different regions within an organism, such as simulating the progression of a signaling cascade or the spatial spread of a cellular response. 

\textbf{Inter-Domain Operations:} These operations facilitate cross-scale manipulation and experimentation by bridging different levels of biological organization. For instance, modeling interactions between transcription factors and promoters can elucidate gene regulatory networks. Simulating genetic perturbations at the cellular level can help predict the effects of gene knockouts or overexpression on cell function. Co-modeling host-pathogen dynamics allows for the study of infection processes and immune responses by simultaneously representing both the pathogen and host cellular environments.

\textbf{Additional Operations:} Multi-resolution scaling enables seamless transitions between different levels of granularity, allowing analyses that span from molecular to organismal scales. Agent-based modeling can simulate the behaviors of individual cells or molecules within a larger system, providing insights into emergent phenomena resulting from complex interactions. Such operations support the exploration of biological processes like tissue development, immune responses, and population dynamics.

By integrating these computational operations, an AIDO becomes a powerful framework for simulating and understanding the multifaceted nature of biological systems. It allows researchers to manipulate and analyze biological entities {\it in silico} across various scales and contexts, ultimately facilitating advances in biomedical research and personalized medicine.

\subsection{Predictive, Generative, Programmable Biology}

Once we have built such an AI-driven digital organism, equipped with the rich set of operations described above, we envision utilizing it for a variety of tasks across multiple biological scales:

• \textbf{Predicting biological phenomena at all levels}: Examples include inferring protein structures from amino acid sequences, determining phenotypes from genotypes, and forecasting cellular responses to specific perturbations.

• \textbf{Simulating biological processes comprehensively}: This includes modeling genetic manipulations, molecular designs, drug effects, and treatment outcomes to understand their impact on biological systems.

• \textbf{Experimenting with perturbations and interventions}: An AIDO allows for virtual experimentation across different scales, enabling us to observe the effects of various genetic or environmental changes without physical limitations.

• \textbf{Programming biological functions and circuitry}: By designing and testing functions of new biological molecules and regulatory networks with specific characteristics, such as increased therapeutic efficacy or reduced side effects, we can advance synthetic biology and develop novel therapeutic strategies.

• \textbf{Evolving biological systems in silico}: Subjecting virtual molecules or systems to simulated fitness landscapes or adversarial selection pressures can help us study evolutionary processes and optimize biological functions.

Such a system enables the generation and manipulation of biological content at all levels—including gene sequences, regulatory programs, protein structures, network topologies, cell designs, tissue architectures, and modeling organism phenotypes —thus providing a virtual platform for comprehensive biological experimentation and design.

\subsection{Why an AI-driven Digital Organism is Superior} 

All biological entities form a holistic system encompassing multiple scales, from molecules to organisms to ecosystems. Biological problems are inherently interconnected rather than isolated and can be studied and addressed at various nested levels of granularity. Although advancements in experimental technologies offer more data, conducting biological experiments remains expensive and time-consuming. Given the exponential growth of historical data, there is an urgent need to model and capture the information within these datasets, leveraging the generalization capabilities of computational models to extract more value from existing data, and reduce the need for a large amount of costly wet-lab experimental data. 

By building an AIDO, we can create a digital experimental environment that is more affordable, safer, faster, programmable, repurposable, and highly adaptable to multiple tasks. Conducting experiments, designing, and programming biology within such a digital environment—and then selectively iterating, refining and improving the AIDO with wet-lab experimental approaches—can significantly reduce the reliance on physical experiments and accelerate innovation in the biological sciences.

\begin{figure}
    \centering
    \includegraphics[width=0.98\linewidth]{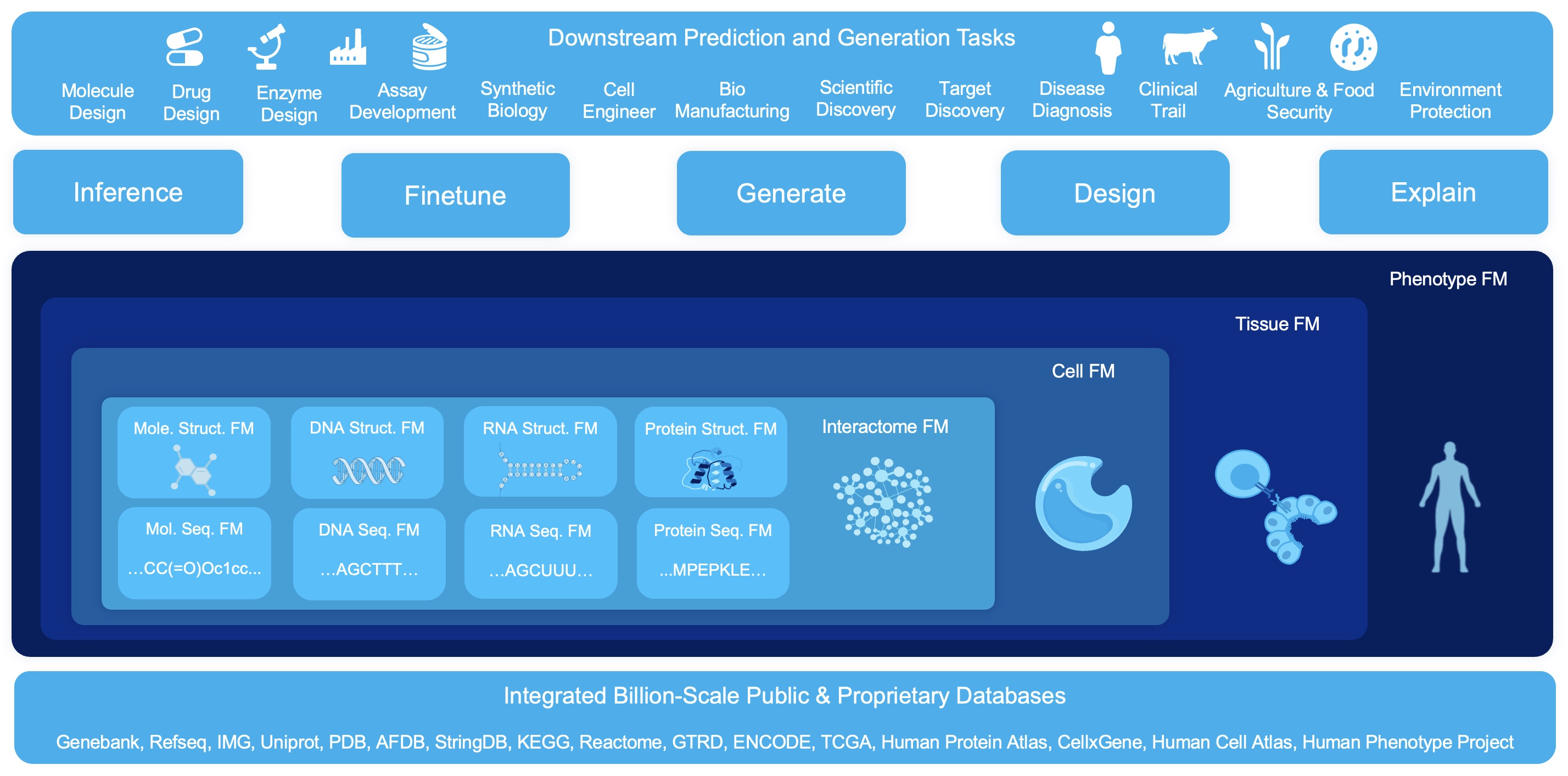}
    \caption{An AI-driven digital organism (AIDO) is a system of multiscale foundation models for biology, which consists of 4 layers: data layer, foundation model system layer, downstream utility layer, and applications of various types of bioengineering problems.}
    \label{fig:multiscalefm4bio}
\end{figure}

\section{How to Build an AI-driven Digital Organism}

In the following, we present our perspective on constructing an AIDO. To accurately represent the multiscale and interconnected nature of biological systems, foundation models for biology must reflect these complexities. We posit that an AIDO shall be built in a modular and connectable way such that these modules can be combined and cascaded to model and address problems arising at different biological scales and complexities (See Figure~\ref{fig:multiscalefm4bio}). Furthermore, the development of these foundation models shall consider the substantial amount of existing data available in the field and anticipate the increasing influx of data in the near future. Taking into account these aspects, an engineering viable approach to building the AIDO is to develop it in 3 stages, i.e., the module building stage, the module connection stage, and the system unification stage. In the following, we will expose the principle we use to build the AIDO, the data available for such development, and the concrete work essential for different stages of the development. 

\subsection{Foundation Model Paradigm}

Classical machine learning predictive models are typically trained on labeled datasets specific to a particular task. However, their accuracy is often limited by the scarcity of labeled data, and they generally exhibit poor transferability to other tasks. In contrast, the paradigm of foundation models~\cite{bommasani2021opportunities} involves pretraining models on large amounts of unlabeled data using self-supervised objectives such as masked language modeling (MLM)~\cite{kenton2019bert}, next token prediction (NTP or GPT)~\cite{radford2018improving}, auto-encoding (AE)~\cite{van2017neural}, and contrastive learning (CL)~\cite{oord2018representation,radford2021learning}. In MLM and NTP, models learn to recover parts of the input that are intentionally hidden, while in AE, they reconstruct the entire input from compressed latent representations. Contrastive learning trains models to produce similar embeddings for similar inputs while distinguishing between different ones.

Architecturally, pretrained models often employ transformer architectures, which utilize pairwise attention mechanisms to capture long-range interactions within input data~\cite{vaswani2017attention}. State-space models (SSMs), such as Mamba~\cite{gu2023mamba}, which use recurrent architectures to capture long-range dependencies are also often used in constructing pretrained models. Beyond sequential data, graph neural networks (GNNs)~\cite{kipf2016semi}, message passing neural networks~\cite{gilmer2017neural} and geometric deep learning (GDL)~\cite{bronstein2017geometric} are also employed to model more complex input structures or dependencies which are represented as graphs. In GNNs or GDL, message-passing operations propagate information through the graph, producing node and edge representations after several iterations. Recently, diffusion models have also been utilized for continuous and discrete outputs to build foundational generative models or decoders for various types of geometric and sequence data~\cite{watson2023novo,ingraham2023illuminating,alamdari2023protein}. These models can be pretrained on existing structure and sequence datasets to model the distribution of input data and later adapted for specific prediction tasks or conditioned for specific generation tasks.

While recent foundation models are becoming multimodal—exemplified by models like GPT-4~\cite{achiam2023gpt} and Gemini~\cite{team2023gemini}—the multimodal data used for cognitive or world modeling fundamentally differs from that in biological modeling. Biological systems “speak” a language distinct from human language. For instance, the relationships among the three primary biological modalities—DNA, RNA, and proteins—are governed by the central dogma of molecular biology, which significantly differs from relationships in multimedia data. These modalities exhibit high levels of redundancy, alignment, and cascading information logic. DNA influences RNA and protein function not only through sequence-defined structures but also via expression levels, temporal dynamics, post-translational modifications, spatial contexts, and co-expression patterns. Furthermore, many of the causal logic and mechanisms in biological systems remain unknown, including the temporal, spatial, and cell-specific behaviors of gene products. 

Consequently, large language models built on human texts and internet images are not directly applicable, and a new set of foundation models is needed. Furthermore, developing foundation models for biology requires new architectures with appropriate tokenization, context lengths, specialized attention mechanisms, latent representations, hierarchical structures, and calibration tailored to biological data. These models must account for the unique characteristics of biological information, including its multiscale organization and the complex interplay between different biological entities and processes, which we address with concrete expositions in this paper.

\subsection{Available Data}

Training the foundation model components that constitute an AIDO necessitates vast amounts of data encompassing biological scales, and thus the types of models we can develop are intrinsically linked to data availability. The continuous reduction in sequencing costs and the advent of high-throughput experimental methods have led to a rapid increase in datasets suitable for self-supervised learning in biology.

\textbf{Biological sequences.} Major repositories such as the National Center for Biotechnology Information (\textbf{NCBI}\footnote{\url{https://www.ncbi.nlm.nih.gov/}}), the European Bioinformatics Institute (\textbf{EBI}\footnote{\url{https://www.ebi.ac.uk/}}), the DNA Data Bank of Japan (\textbf{DDBJ}\footnote{\url{https://www.ddbj.nig.ac.jp/index-e.html}}), and the Integrated Microbial Genomes and Microbiomes system hosted by the Joint Genome Institute (\textbf{JGI}\footnote{\url{https://img.jgi.doe.gov/}}) house extensive sequencing data from a wide range of species, including vertebrates, invertebrates, and bacteria. Specialized databases have emerged from these repositories, such as \textbf{Ensembl}\footnote{\url{http://www.ensembl.org/}} for meticulously annotated genomes, \textbf{UniProt}\footnote{\url{https://www.uniprot.org/}} for protein sequences, and \textbf{RNAcentral}\footnote{\url{https://rnacentral.org/}} for noncoding RNAs. For instance, there are tens of thousands of complete genomes, each comprising billions of nucleotides. Additionally, hundreds of millions of noncoding RNAs and billions of proteins have been sequenced.

\textbf{Molecular structures.} The Protein Data Bank (\textbf{PDB}\footnote{\url{https://www.rcsb.org/}}) contains over 200,000 entries of proteins and other molecular complexes. Moreover, more than 10,000 RNA structures have been documented. Predicted structures with associated confidence levels are also abundant, thanks to resources like AlphaFold Protein Structure Database (\textbf{AlphaFold DB}\footnote{\url{https://alphafold.ebi.ac.uk/}}) and ESM Metagenomic Atlas (\textbf{ESM Atlas}\footnote{\url{https://esmatlas.com/}}), which collectively provide hundreds of millions of predicted protein structures. Recently, there are also increasing amount of DNA packing structure data available allowing us to study the structural organization of genomes in relation to their regulatory roles. 

\textbf{Interactome and relational data.} In the realm of molecular interactions, databases such as \textbf{STRING}\footnote{\url{https://string-db.org/}} (Search Tool for the Retrieval of Interacting Genes/Proteins) offer more than 10 billion physical or inferred relationships between molecules. Pathway databases like \textbf{KEGG}\footnote{\url{https://www.genome.jp/kegg/}} (Kyoto Encyclopedia of Genes and Genomes), \textbf{Reactome}\footnote{\url{https://reactome.org/}} and \textbf{BioCyc}\footnote{\url{https://biocyc.org/}} provide detailed maps of biological pathways, while the Gene Transcription Regulation Database (\textbf{GTRD}\footnote{\url{https://gtrd.biouml.org/}}) focuses on transcription factors regulating gene expression. Gene Ontology (\textbf{GO}\footnote{\url{https://gtrd.biouml.org/}}) also defines the relation between the genes and the biological pathways in a hierarchical fashion.  

\textbf{Transcriptome and cellular activity.} Single-cell RNA sequencing (scRNA-seq) and spatial transcriptomics represent rapidly growing data sources. Public repositories now contain over 100 million scRNA-seq measurements and thousands of spatial transcriptomic datasets. Consolidated platforms like the Chan Zuckerberg Initiative’s \textbf{CellxGene}\footnote{\url{https://cellxgene.cziscience.com/}} facilitate access to and analysis of these datasets. There is also an increasing amount of experimental data measuring cellular response to perturbations, such as the Genomics of Drug Sensitivity in Cancer (\textbf{GDSC}\footnote{\url{https://www.cancerrxgene.org/}}) dataset.  Beyond gene expression data, there is an expanding wealth of spatial information on the transcriptome, including imaging data for cells and tissues. Projects like the Human Protein Atlas (\textbf{HPA}\footnote{\url{https://www.proteinatlas.org/}}), Human Cell Atlas (\textbf{HCA}\footnote{\url{https://www.humancellatlas.org/}}), and Jump Cell Painting dataset (\textbf{Jump-CP}\footnote{\url{https://jump-cellpainting.broadinstitute.org/}}) are generating extensive imaging datasets that provide spatial context to transcriptomes. \textbf{Cell Ontology}\footnote{\url{https://obofoundry.org/ontology/cl.html}} is also constructed to provide a structured controlled vocabulary for cell types in animals.

\textbf{Phenome and population data.} Expanding beyond cellular and tissue transcriptomes, large human cohorts are emerging that encompass simultaneous measurements across multiple biological scales and modalities in human subjects. Notable examples include the \textbf{UK Biobank}\footnote{\url{https://www.ukbiobank.ac.uk/}}, The Cancer Genome Atlas (\textbf{TCGA}\footnote{\url{https://www.cancer.gov/ccg/research/genome-sequencing/tcga}}), and the Human Phenotype Project (\textbf{HPP}\footnote{\url{https://humanphenotypeproject.org/home}}). The HPP, in particular, provides broad clinical, physiological, genetic, transcriptomic, cellular, and phenotypic data measurements over time, offering a rich resource to model complex dependencies and link fine-scale molecular data with higher-level phenotypic outcomes.

These extensive datasets lay the groundwork for building large-scale foundation models across all scales of biology. The pretrained models resulting from these datasets can be employed individually or integrated to address a wide spectrum of biological and life science challenges, from molecular design and cellular engineering to systems biology and personalized medicine.

\subsection{Multiscale Foundation Models for Biology}

In the following, we outline a technical blueprint to develop a system of foundational models capable of addressing biological questions across different scales. The overall scheme consists of 3 stages. The first stage is to build up a necessary set of fundamental building blocks or modules representing the major data modalities arising in biology -- in a ``divide-and-conquer" fashion. The second stage is to develop a set of new deep learning architectures that integrate the central dogma, regulatory rules, and the interconnected nature of biology, as well as different data modalities or modules in a bottom-up fashion to reflect the multiscale, nested, and hierarchical organization of biological systems. These architectures can bridge the existing gaps by integrating biological knowledge into the models and developing models that can seamlessly operate across various biological scales and modalities -- thereby ``dots are connected". In the third stage, the modules and connected modules are unified into a networked system, where representations and embeddings can be passed around in different nodes and levels of the systems, and especially feedback and gradient signals from the coarser and topper level of the system can be propagated all the way back to the bottom level of the system to further improve these modules. This is like the ``aligning and optimization" phase in an assembly process. With a set of benchmarks and supervisory tasks from different levels and scales of biology, all the system modules can be jointly adapted and aligned to achieve synergy towards an overall better or even emergent system-level performance.  

\subsubsection{Divide-and-Conquer: Models for Specific Modalities and Scales} 

\begin{figure}
    \centering
    \includegraphics[width=0.98\textwidth]{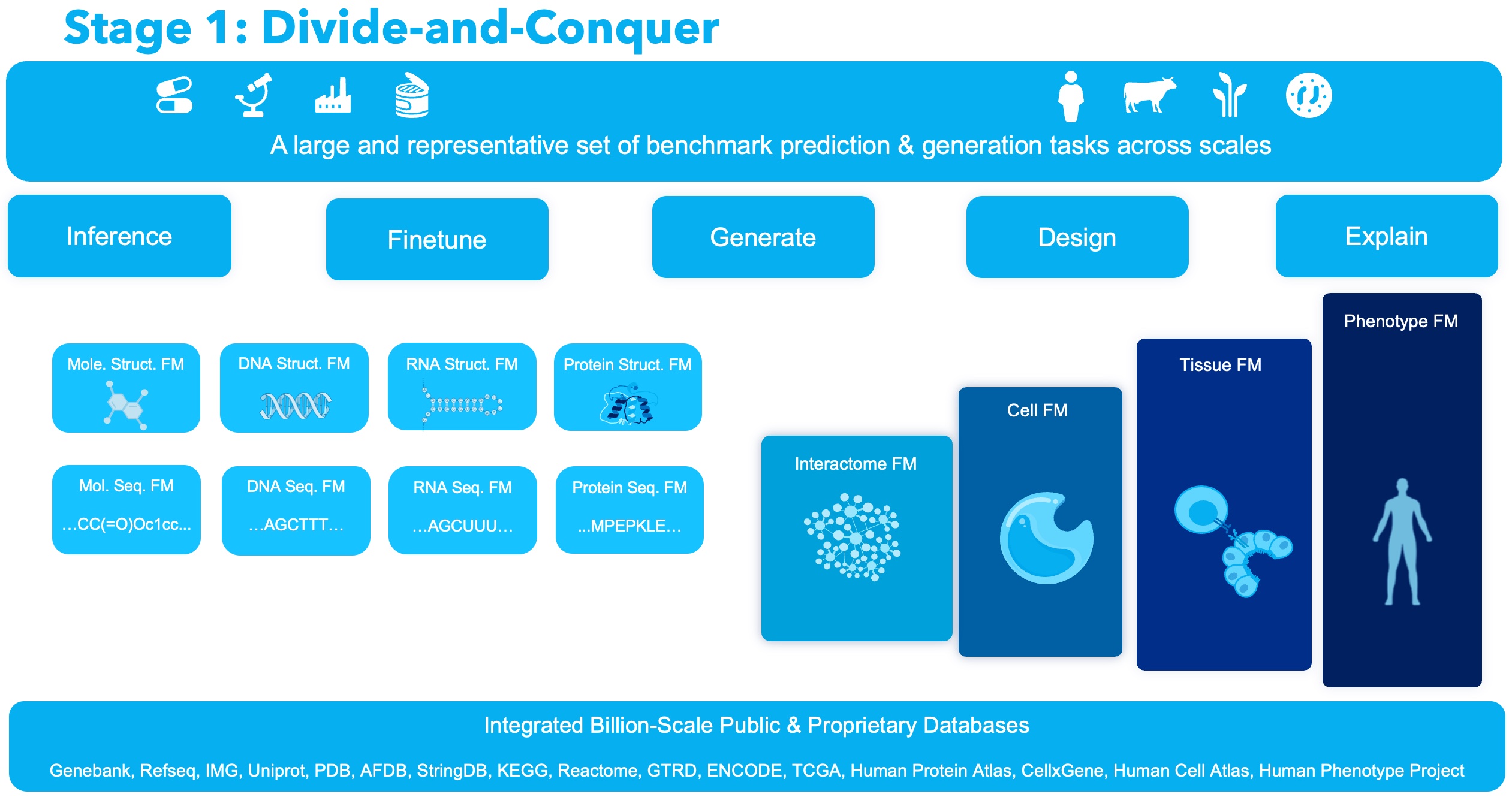}
    \caption{Stage 1 of building an AIDO.}
    \label{fig:mulstiscalefm_stage1}
\end{figure}

Pretrained models for each data modality are important building blocks for more complex models. These initial models can also be continuously pretrained with data from specific domain and evolved in a lineage into new pretrained models. Therefore, the initial step towards an AIDO is to build a family of models which can be used independently and re-used as building blocks for more advanced models and integrated systems (See Figure~\ref{fig:mulstiscalefm_stage1} for a summary). In the process of building these modules, new tokenizers of the biological data and new architectures also need to be developed to better process the input biological data and learn better representations for these data.    

\subsubsection*{Individual Pretrained Models for Each Data Modality}

We will first pretrain both sequence and structure models for molecule-level data involved in biology, such as small organic molecules, DNA, RNA and protein sequences. Next, graph neural networks can be used to pretrain regulatory networks and pathway data from KEGG, Reactome and GTRD. Single-cell pretrained models can be constructed based on data from cellxgene, and tissue and cell type image models can be pretrained over human cell atlas and human protein atlas data. Lastly, some of the phenotype pretrained models, such as sleep, gait, electrocardiograms (ECG), and glucose response models, can be obtained from data collected in the Human Phenotype Project. 

In recent years, there has been a notable surge in efforts within academia and industry to explore the opportunities presented by pretrained large models for one of these modules listed above. For DNA sequences, models like Nucleotide Transformer~\cite{dalla2023nucleotide}, HyenaDNA~\cite{nguyen2024hyenadna} and Evo~\cite{nguyen2024sequence} have been developed, leading to improvements in a range of genome-related downstream tasks. Similarly, RNA-FM~\cite{chen2022interpretable} and CodonBERT~\cite{li2023codonbert} have been created for RNA analysis. In the realm of protein sequences, models such as ESM~\cite{lin2023evolutionary}, ProGen2~\cite{nijkamp2023progen2}, and xTrimoPGLM~\cite{chen2024xtrimopglm} have been introduced. Advances in protein structure prediction have been achieved with models like AlphaFold~\cite{jumper2021highly,abramson2024accurate}, ESMFold~\cite{lin2023evolutionary}, and GearNet~\cite{zhang2022protein}. For single-cell RNA sequencing, models such as GeneFormer~\cite{theodoris2023transfer}, scFoundation~\cite{hao2024large}, and scGPT~\cite{cui2024scgpt} have been developed. Additionally, BioGPT~\cite{luo2022biogpt} and MedSAM~\cite{ma2024segment} have been applied to biomedical text documents and images. The amount of data available for pretraining these models has grown beyond billions of data points, and model sizes have been scaled to internet-scale levels with parameters exceeding 100 billion~\cite{chen2024xtrimopglm}. However, most of these pretrained models leverage transformer architecture and are trained via MLM or GPT objective and applied to raw biological inputs. We will next explain the need and the design of new tokenizers for biological data and new architectures learned from these data. 

\subsubsection*{New Tokenizers for Biological Data}

New tokenizers also need to be developed for biological data due to their diversity and differences from natural language and images.

\textbf{Tokenization for biological sequences.} For DNA and RNA sequences, every three nucleotide bases in the coding region correspond to an amino acid, but for noncoding region, such correspondence does not exist. Thus, in order to take into account such biological information, tokenizers for DNA and RNA sequences will need to be designed in a way which aligns with the boundary of coding and noncoding regions. Furthermore, for coding regions, a vocabulary for a combination of a group of three or multiple of three nucleotides needs to be used, but for noncoding regions, there is no such restriction on aligning with the reading framework of the coding regions. Therefore, for the noncoding region, we can use a more information-driven approach to tokenize the sequences, such as the Byte-Pair Encoding (BPE) tokenizer~\cite{sennrich-etal-2016-neural} which can come up with variable length tokens based on sequence statistics. Thus, for biological sequences, such hybrid tokenizers seem to be needed to best represent the input by taking into account both information theory and biological knowledge. 

\begin{figure}
    \centering
    \includegraphics[width=0.65\textwidth]{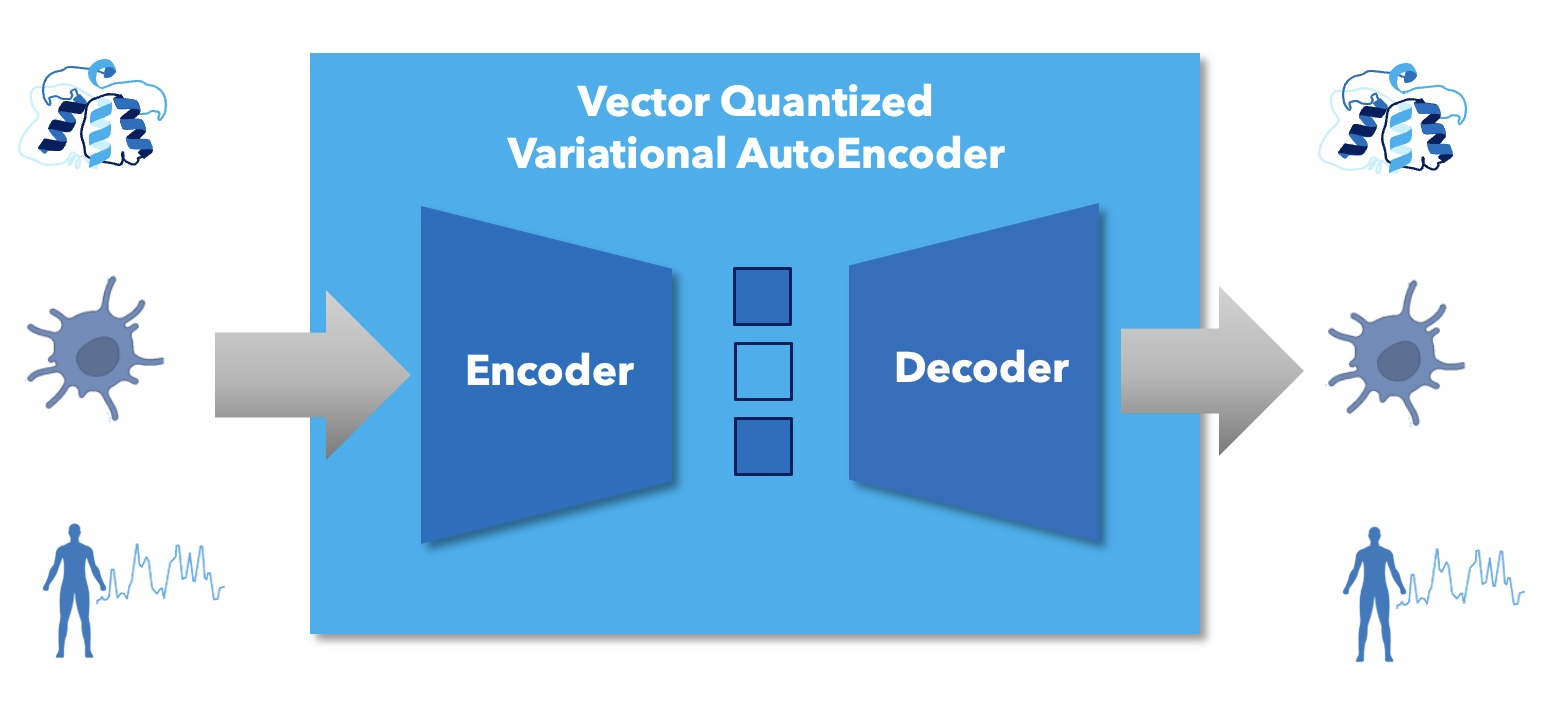}
    \caption{Tokenizer for complex biological data which leverages vector quantized variational autoencoder architecture. Tokenized or quantized biological data can then inter-operate better with other discrete information and operations in foundation models.}
    \label{fig:vqvae_tokenizer}
\end{figure}

\textbf{Tokenization for molecular structures.} 
Developing accurate models for molecular structures and properties at the molecular level is of paramount importance. 
To model the 3D structures of biomolecules and their interactions, which is pivotal for understanding their functions and guiding the design of novel therapeutic agents, more fine-grained and informative tokenization of structural input is essential. 
With the extensive availability of experimentally determined and computationally predicted structures in repositories such as the Protein Data Bank (PDB), the AlphaFold Protein Structure Database (AFDB), and the ESM Atlas, there is an unprecedented opportunity to leverage these data using advanced computational techniques. We propose employing geometric deep learning models as encoder and vector quantized variational autoencoders (VQ-VAEs), to learn compact representations of biomolecular structures~\cite{van2024fast,zhang2024balance,hayes2024simulating} (also see Figure~\ref{fig:vqvae_tokenizer}). By encoding the intricate 3D conformations into discrete tokens within a latent space, the model captures essential structural features while achieving significant data compression. A complementary deep learning structure decoder model can then reconstruct the full 3D structures from these compressed representations. 

Training the structure decoder using diffusion models further enhances this approach by enabling the generation of an ensemble of molecular conformations, thereby capturing the inherent flexibility and dynamics of biomolecules. This probabilistic modeling of structural variability is crucial for understanding functional mechanisms and for designing molecules with desired properties.

This compression-decompression framework offers several advantages. First, the encoder learns critical features of the input structures, which can be harnessed for downstream applications such as structure retrieval, molecular docking, and inverse folding problems. Second, the discrete latent representations, or structure tokens, can be integrated into protein language models, enriching them with structural context and improving their predictive capabilities. Third, by aligning amino acid sequences with corresponding structure token sequences—made possible by the parallel data in PDB, AFDB, and the ESM Atlas—we facilitate the development of models that can predict structural information directly from sequence data. This alignment between sequence and structure allows for the prediction of structure token sequences from amino acid sequences, effectively bridging the gap between primary sequence information and 3D structural conformations. Utilizing the structure decoder, we can reconstruct detailed molecular structures from these predictions, creating a powerful tool for structure prediction and validation~\cite{zhang2024balance}. The modular design of the encoder and decoder not only simplifies the integration with other computational modules but also enhances the scalability and adaptability of the model. This flexibility enables the extension of the framework to accommodate various types of biomolecules and complexes, broadening its applicability. 

\textbf{Tokenization for cellular transcriptome and image data.} 
A large amount of single cell gene expression and other transcriptomic data are measured and curated nowadays, and these measurements reflect the continuous concept of how much each biomolecule is present in a dynamical system. To better integrate these continuous measurements into a unified system and allow for inter-modality operability, we will need to learn to embed or tokenize these inputs. Furthermore, increasingly, high-content imaging data that reveal cellular morphology are being collected alongside transcriptomic data, and adding spatial information about the location of biomolecules. To incorporate morphological and location features into our modeling framework, these information also need to be embedded or tokenized to allow for unified modeling and cross-operability. We can employ again autoencoder or vector quantized autoencoder architectures to learn latent representations of transcriptomic and cellular image data~\cite{palma2023predicting} (See Figure~\ref{fig:vqvae_tokenizer}). Since we need to handle both individual molecule representation and cellular representation from such data, new architecture and training methods will be needed in the VQ-VAE framework to accommodate such two levels of tokenization and to ensure consistency between the two levels of tokens. Furthermore, due to the introduction of spatial information in imaging and spatial transcriptomic data, such spatial information also needs to be encoded and attended to during the representation learning process.  

\textbf{Tokenization for phenotype information.} Beyond single-cell and spatial transcriptomics, modeling tasks at the organ, organ network, and individual levels are essential for a comprehensive understanding of complex biological systems and diseases. The availability of data at these higher scales is rapidly increasing, thanks to large-scale population cohorts such as the UK Biobank and the Human Phenotype Project (HPP). These resources provide extensive phenotypic data, including molecular, longitudinal and time-series measurements, which can be leveraged to build robust models given sufficiently large and standardized cohorts.

Complex time series of phenotypic measurements—such as continuous glucose monitoring patterns, sleep quality metrics from wearable devices, dietary intake logs, and physical activity levels tracked by accelerometers—can also be modeled using a continuous or vector-quantized autoencoder framework with transformers and other deep learning encoders specialized for phenotype time-series data. These models can learn latent continuous or discrete representations that capture essential features influencing health and disease states (See Figure~\ref{fig:vqvae_tokenizer}). We can also employ contrastive learning objectives together with reconstruction objectives to enhance the robustness of the representations~\cite{levine2024genetic}. 

\subsubsection*{New Deep Architectures beyond Transformers} 

Biological data and the mechanisms behind them pose unique modeling challenges not seen in common data such as natural texts and images. New deep learning architectures need to be designed to 
handle many properties unique to biology data and mechanisms. 
For instance, protein structures consist of atoms situated in 3D spaces, and the representations of 3D structures need to be invariant to the translation and rotation. 
For example, biological sequences such as DNA are not only extremely long but also behold long-range interactions at various granularity, necessitating modeling of such distant input contexts. For instance, to model a eukaryotic gene, we need to take into account both introns and exons as well as regulatory regions such as promoters and enhancers, which may involve hundreds of kilobases. Another example is in modeling human single cell gene expression data, where around 20k dimensional count data need to be modeled simultaneously to capture the cell states. The underlying regulatory mechanisms involving genome sequences, RNA, and proteins make such data complex and challenging to model.  
Standard deep learning architectures, such as conventional transformers, face challenges either due to such architectures not expressing the proper inductive bias or due to computational and memory constraints when processing such complex biological contexts. Therefore, designing architectures capable of handling unique biological inputs and contexts in a memory- and computationally efficient manner is essential. 

\textbf{Architectures for molecular structures.} For molecular structures with 3D coordinate information, we need to incorporate some of the physical constraints in the representation learning in different applications, such as equivalences, where embeddings are rotated and shifted as these operations are applied to input structures, and invariances, where embeddings remain the same irrespective of the rotation and shift of the input structures. To be able to systematically address these physical constraints, we will need to design deep architectures in the spherical harmonic space where embeddings are learned as vector spaces of
irreducible representations and have sparse message passing or dense attentions between nodes based on equivariant operations such as tensor products~\cite{liao2022equiformer}. However, a bottleneck in scaling up these deep architectures is the computational complexity
of the tensor products when we use a high number of basis in spherical harmonics. Thus, efficient implementations of these architectures are also needed to make them efficient yet expressive.    

\textbf{Architectures for long sequence inputs.} A family of sparse and hybrid deep architectures has the potential to address the long sequence problem arising from genomic and cellular modeling. For instance, transformers with random sparse attentions~\cite{child2019generating},  hierarchically designed attentions~\cite{ding2023longnet} and the mixture of experts~\cite{jiang2024mixtral,sun2024mixture}, can allow transformers to scale to very long input sequences. These efficient layers can also be stacked many times in a deep architecture allowing information from different parts of the input to sufficiently mix without losing too much of the representation power of the model. As for hybrid architectures, for example, integrating convolutional neural networks (CNNs) with transformers or employing state-space models like Mamba~\cite{gu2023mamba} offers significant advantages. CNN architectures are proficient at capturing local sequence features, while transformers and SSMs excel at modeling extensive global interactions. By combining these architectures into a hybrid model, we can maintain high representational capacity while achieving greater computational efficiency for processing long sequences and contexts. This hybrid approach allows the foundation model to leverage the strengths of each architecture: CNNs efficiently extract local patterns and motifs within sequences, which are critical for understanding functional domains and active sites, while transformers and SSMs handle the global context and long-range dependencies that are crucial for capturing distant interactions and evolutionary information within proteins or nucleic acids. By adopting such architectures, we can develop models that are better suited to the complex and hierarchical nature of biological sequences, ultimately enhancing our ability to predict structural and functional aspects of biomolecules based on sequence data. 

\textbf{Architectures for high dimensional cellular data.} Single cell RNA-seq data are typically very sparse in the sense that only about 10\% of the genes are measured for each cell and the majority of genes are not measured either due to low expression or technical limitations. Thus to model the gene expression level of about 20k protein coding genes in human cells, a new architecture needs to be designed to balance the expressiveness and efficiency of the architecture. For instance, to differentially deal with expressed genes and non-expressed genes, we can design an asymmetrical encoder-decoder framework specifically for sparse gene expression matrices, which is achieved by feeding only the unmasked non-zero positions (less than 10\% of the full length) into the encoder, while the largely masked and zero positions are input into a lightweight decoder with a reduced number of layers and attention heads~\cite{hao2024large}. Such unique architecture significantly reduces computational
costs and training time. In addition, a novel auto-discretization strategy can also be designed to project continuous expression
values into a latent embedding space. Instead of rounding to the nearest integer, values are directly mapped to the latent
space allowing for the representation of closely related values. 

\subsubsection{Connect the Dots: Integration across Modalities and Scales}

\begin{figure}
    \centering
    \includegraphics[width=0.98\textwidth]{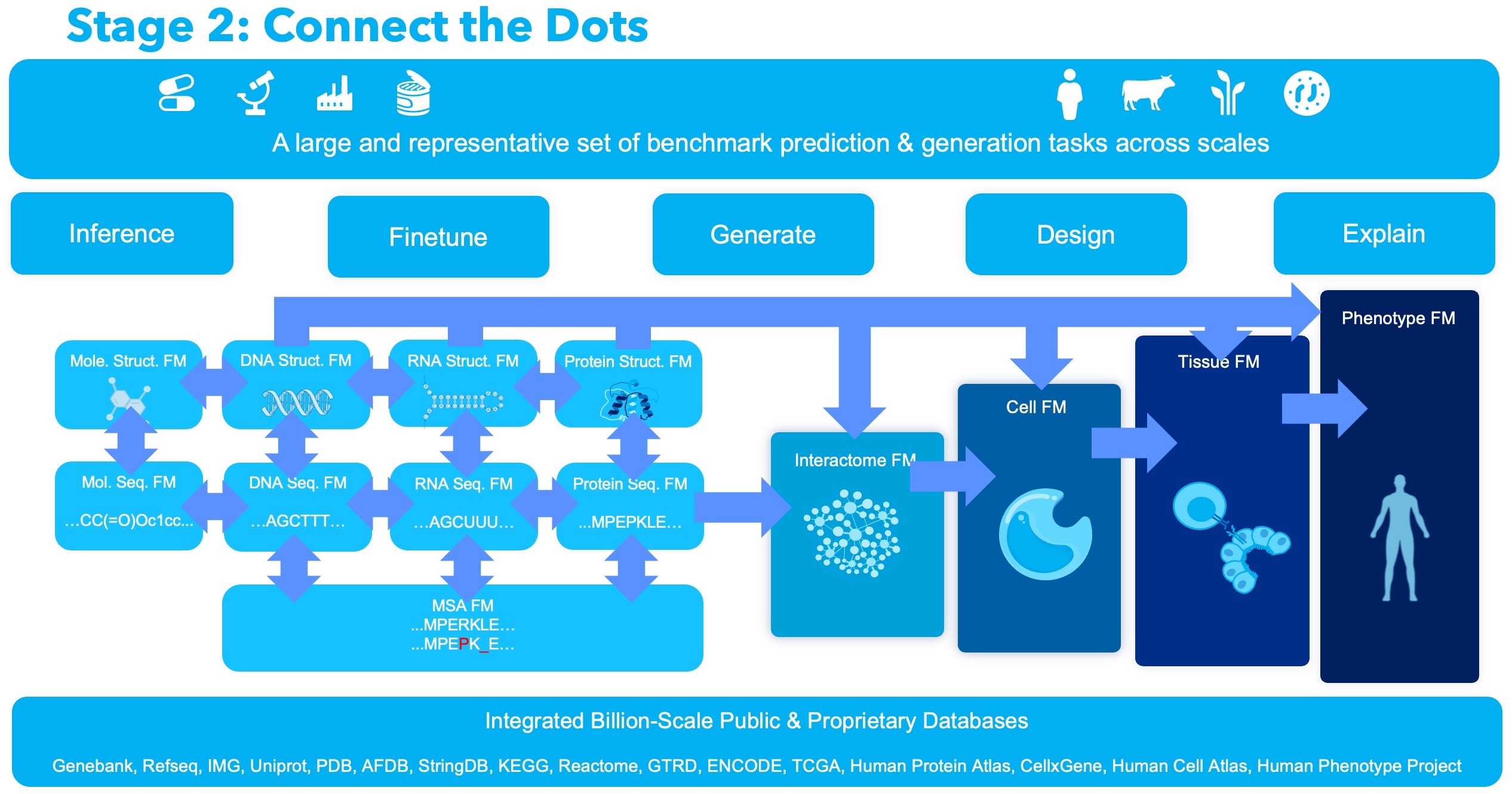}
    \caption{Stage 2 of building an AIDO.}
    \label{fig:mulstiscalefm_stage2}
\end{figure}

Once we have the modules for different modalities and scales of biological data, we can connect and combine these modules to address more complex biological problems, and build better models by linking information arising from different scales of biology (See Figure~\ref{fig:mulstiscalefm_stage2} for a summary). We will use several technical approaches that are reusable for connecting different modules for more advanced modeling.

\subsubsection*{Markup Language Models for Integrating DNA, RNA, and Protein Sequences}

Recent large language models (LLMs) for biological sequences are developed separately for each type of molecule—DNA, RNA, or proteins. While this specialization has led to advances in modeling each modality, it overlooks the fundamental interconnectedness of these entities as described by the central dogma of molecular biology. DNA encodes the information for RNA and proteins, and there is a direct correspondence between these sequences. This presents an opportunity to unify these three types of sequences within a single foundation model that leverages their intrinsic relationships. 
A recent attempt to bridge this gap is the Evo model, which utilized a pretrained DNA model to address downstream tasks involving RNA and proteins. However, this model was limited by its training data, which consisted solely of bacterial genomes, and it did not incorporate crucial biological annotations such as regulatory regions, noncoding RNA regions, or coding regions with introns and exons. Moreover, while the number of complete reference genomes is limited, there exists a vast amount of data on sequenced and expressed RNA and proteins from incomplete genomes.

\begin{figure}
    \centering
    \includegraphics[width=0.95\textwidth]{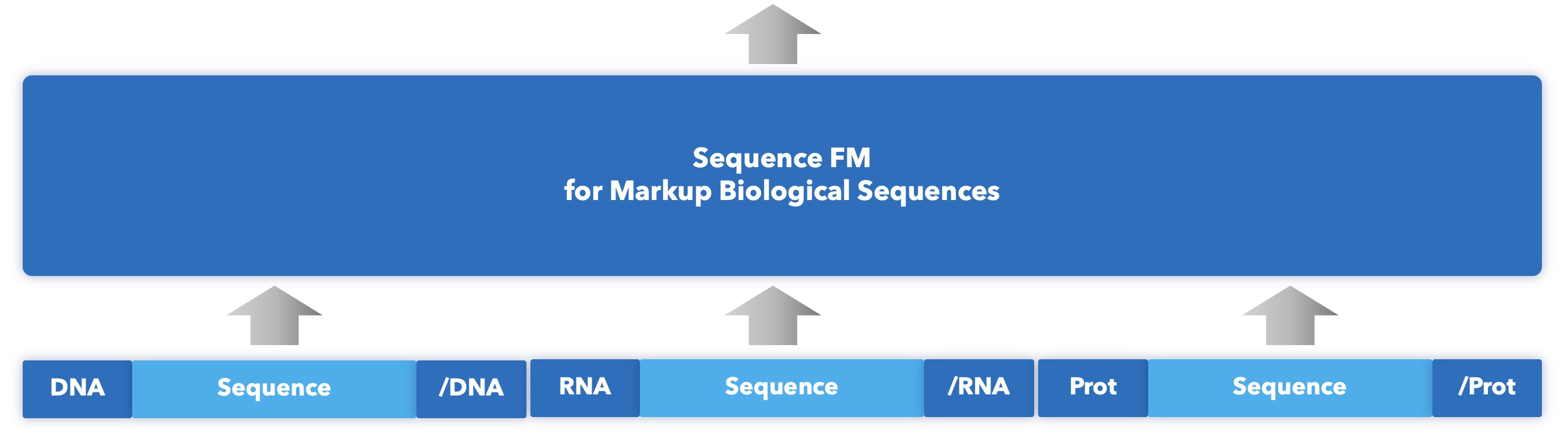}
    \caption{FM for central dogma which leverages markup information to unify DNA, RNA and protein data with different levels of completeness.}
    \label{fig:fm4seqmarkup}
\end{figure}

To fully harness the available biological sequence data, one can develop a markup language model for biological sequences (Figure~\ref{fig:fm4seqmarkup}). This model would integrate rich annotation information for different functional units within the genome and leverage fragments of expressed sequences to maximize data utilization. By incorporating annotations directly into the sequence data, we can provide the model with context that is essential for understanding biological functions. In practice, this approach involves augmenting the sequence data with labels that indicate the type of molecule—DNA, RNA, or protein—and potentially more fine-grained information such as regulatory elements or coding regions. For example, specific tokens or markers can be inserted at the beginning and end of sequences to denote their biological context. This additional information effectively conditions the model on the type of sequences it is processing, enabling it to generate more relevant representations and predictions. Furthermore, by unifying DNA, RNA, and protein sequences within a single model and providing type indicators, we facilitate the transfer of information between these modalities. The inherent similarities and correspondences among these sequences can potentially be exploited this way by the model to improve learning and generalization. Such a unified model has the potential to outperform separate models by capturing the holistic nature of genetic information flow and leveraging the vast amounts of available data across all three modalities. This integrated approach aligns with the hierarchical and nested structure of biological systems, reflecting the multiscale organization inherent in biology. By developing foundation models that encapsulate the relationships between DNA, RNA, and proteins, we can create powerful tools for a wide range of downstream tasks, from predicting gene expression and protein folding to understanding regulatory networks and disease mechanisms.

\subsubsection*{Advanced Position Encoding Schemes For Rich Biological Contexts}

Unlike words marked by their unique linear positions in a natural language sentence or sequence, every unit of elements within a biological data collection (e.g., sequence, gene expression) has a myriad of contextual relevance that often overlap, including evolutionary position, chromosomal position, network position, gene-ontology position, cellular position, etc. 

For instance, evolutionary relationships among species result in significant similarities within their genomes; aligning evolutionarily related biological sequences such as genome sequences and protein fragments allows for the identification of conserved regions and regions that fixate more rapidly. These patterns of evolutionary conservation are directly linked to biological functions \cite{u_correlated_1994}. For instance, distant regions in a protein that are conserved or co-evolved may indicate key functional sites essential for maintaining normal biological activities. Thus, multiple sequence alignments (MSAs) and the ability to represent alignment information play a crucial role in genomic analysis and understanding protein functionality. To effectively capture these evolutionary relationships and functional insights within computational models, it is necessary to develop foundation models that can incorporate MSAs in handling biological sequence data. 

Besides multiple sequence alignments, we will need to jointly input sequence and other types of information together in order to come up with better representations. For instance, for proteins, we want to simultaneously input amino acid sequences and tokenized structure sequences to represent a protein. A structure token sequence also has a one-to-one correspondence with its amino acid sequence, requiring us to align them and input them as a 2D matrix to the models. Therefore, it is also necessary to develop foundation models which can handle such aligned data. 

\begin{figure}
    \centering
    \includegraphics[width=0.95\textwidth]{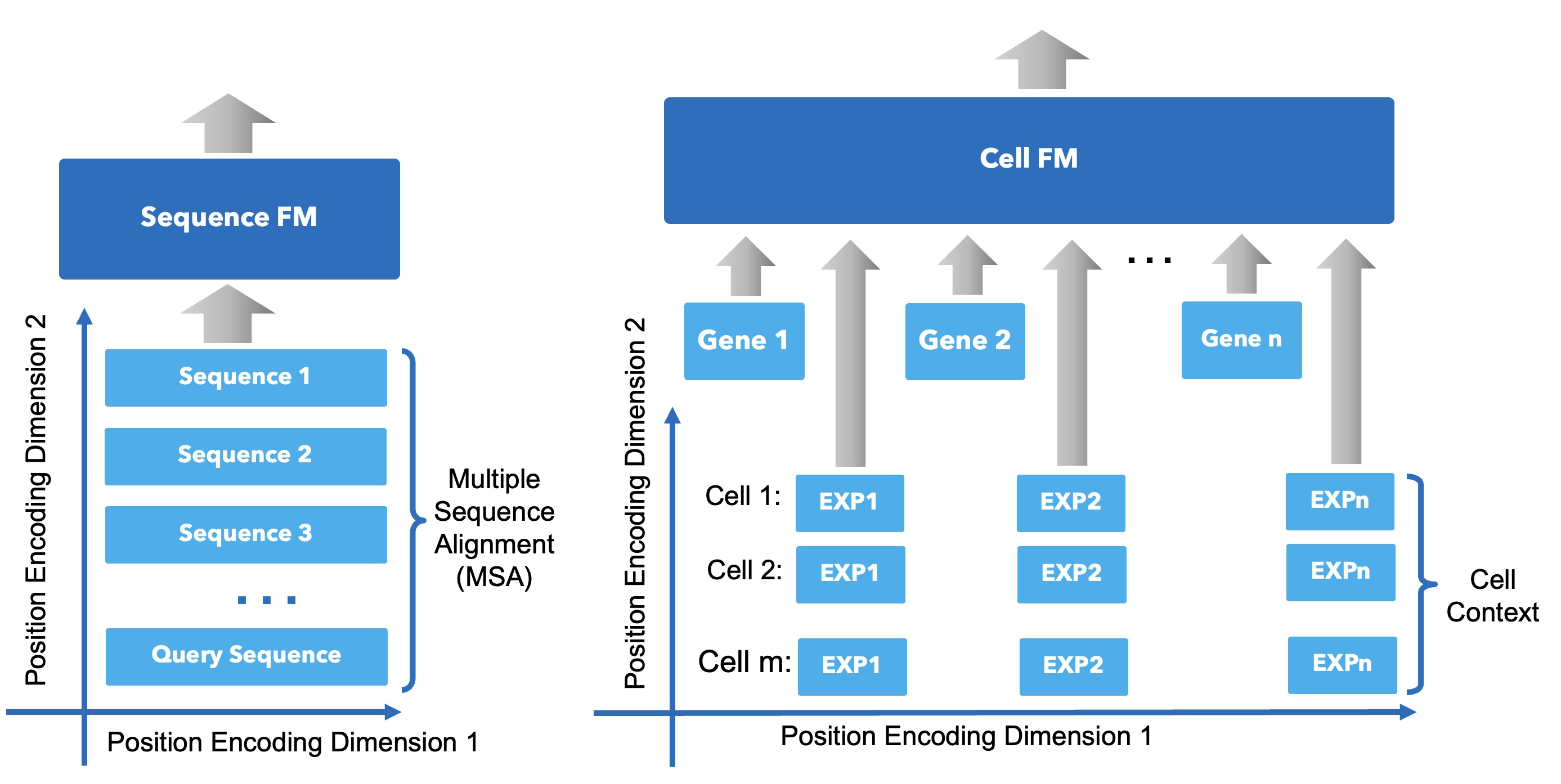}
    \caption{FM with 2-dimensional position encoding for handling rich biological contexts such as multiple sequence alignment and spatial relation between cells}
    \label{fig:fm4position2d}
\end{figure}

One promising approach is to devise new positional encoding methods, such as two-dimensional (2D) positional encoding, which can effectively represent the multidimensional alignment information present in rich biological contexts such as MSAs and multimodal inputs~\cite{chen2024msagpt,sun2024mixture,lee2024ragfold}. Such 2D positional encoding leverages the Rotary Positional Encoding (RoPE)~\cite{su2023enhanced} technique, but specifically adapts it for 2D variant to suit our 2D positional encoding requirements. Conceptually, the 2D positional encoding encapsulates the explicit row- and column-wise self-attention patterns with high efficacy. Moreover, it allows unrestricted information diffusion, that is, enabling any two input tokens to attend to one another. Such a framework facilitates unveiling complex correlation patterns, such as high-order evolutionary correlations among amino acids, that customized self-attentions might overlook (Figure~\ref{fig:fm4position2d}).

The 2D position encoding can also be used to incorporate multi-cellular context into spatial transcriptomic models. Traditional single-cell RNA sequencing (scRNA-seq) models often treat cells as isolated units, ignoring the crucial influence of cell-cell interactions within their spatial environments or micro-environments. In vivo, cells exist within complex tissues where they constantly interact with neighboring cells through direct contact and signaling molecules. These interactions profoundly affect cellular behavior, leading to gene expression profiles that differ significantly from those observed when cells are studied in isolation. Consequently, capturing the spatial context is essential for accurately modeling and understanding cellular functions. To address this limitation, we propose to leverage spatial transcriptomic data to incorporate the gene expression profiles of neighboring cells into the modeling of each target cell. Spatial transcriptomics provides not only the transcriptomic information of individual cells but also their precise spatial locations within the tissue. By identifying the immediate neighbors of each cell, we can construct a more comprehensive representation that reflects both intrinsic gene expression and extrinsic influences from the microenvironment.

Implementing this approach involves extending the positional embeddings used in transformer-based models to include a second spatial dimension (Figure~\ref{fig:fm4position2d}). This method is analogous to the 2D positional encoding employed in transformer architectures for multiple sequence alignments in protein and genomic sequences. Specifically, we encode each center cell along with its spatially adjacent cells, effectively embedding them together to capture spatial dependencies and interactions. By integrating the spatial context through these enhanced positional embeddings, the model can learn patterns that account for the influence of neighboring cells on gene expression. This inclusion of spatial information allows for a more accurate representation of cellular behavior within tissues, improving the model’s predictive capabilities for various downstream tasks. These tasks may include identifying cell types, uncovering spatial domains within tissues, and elucidating intercellular communication pathways. Incorporating multiple-cell context into spatial transcriptomic models not only advances our understanding of cellular function in situ but also has significant implications for fields such as developmental biology, cancer research, and tissue engineering. By capturing the complex interplay between cells within their native environments, we can gain deeper insights into the mechanisms driving tissue organization, disease progression, and responses to therapeutic interventions.

\subsubsection*{Differentiable Computation Graphs for Integrating Pretrained Representations}

Each pretrained module provides an embedding or vector representation of a corresponding type of biological entity, causing diverse biological entities to be projected into one space, offering a basis for vector space operations between them, and enabling them to be combined and cascaded in a nested fashion to form more sophisticated models, reflecting the hierarchical nature of biological entities.

\begin{figure}
    \centering
    \includegraphics[width=0.98\textwidth]{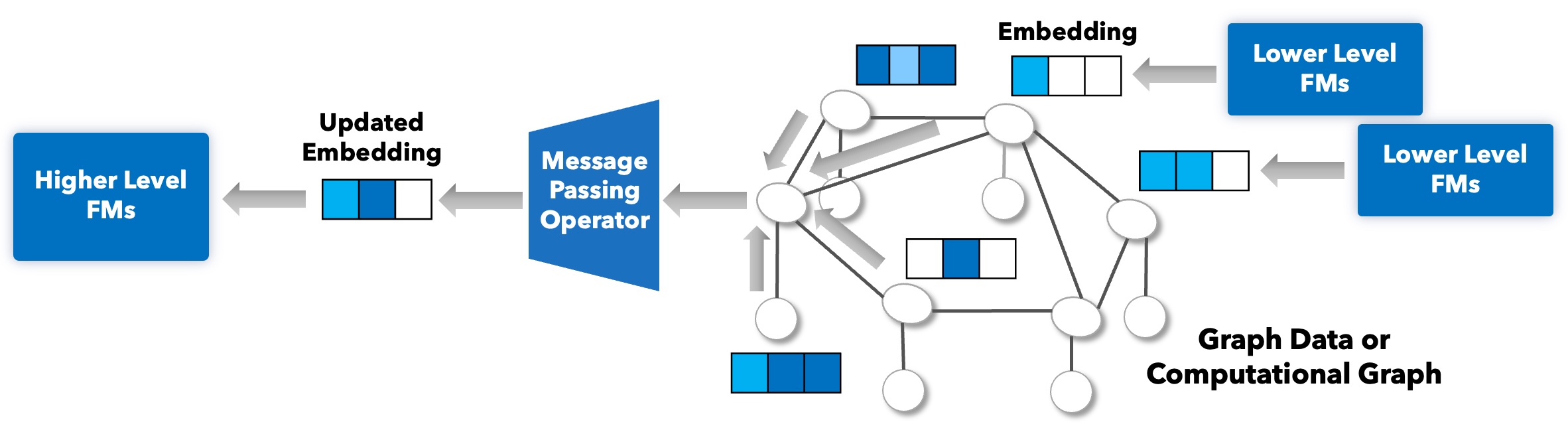}
    \caption{Graph neural networks and more generally differentiable computation graphs can be leveraged to integrate pretrained presentation and build more sophisticated models.}
    \label{fig:gnn}
\end{figure}

Differentiable Computation Graph (DCG) techniques such as graph neural networks (GNNs)~\cite{kipf2016semi} and message passing neural networks (MPNN)~\cite{gilmer2017neural} are well-suited for modeling complex interactions and can accommodate heterogeneous node types and capture complex relations between linked entities and modules, such as the signed effects of edges—representing activation or inhibition genetic or cellular regulations (Figure~\ref{fig:gnn}). More generally, leveraging the differentiable computation ability of modern deep learning platforms, pretrained modules from different levels can also be readily connected into more sophisticated computation graphs allowing embeddings and gradients to be passed between these modules according to the connectivity patterns of the computation graphs.

\textbf{Molecular interaction models on pretrained representations}. 
A substantial body of curated knowledge exists regarding biological pathways and molecular interactions, available through resources such as Reactome and KEGG for pathway information, the Gene Transcription Regulation Database (GTRD) for transcription factor–DNA binding interactions, and STRING for protein–protein interaction networks. These databases provide a rich foundation for pretraining representations of genes and proteins that effectively capture the network effects inherent in biological systems. Leveraging this existing knowledge is particularly crucial when modeling the impact of drug interventions or genetic perturbations. The effects of such perturbations often propagate through molecular networks, influencing downstream entities within biological pathways, and can lead to cascading changes that alter the behavior of entire cells or tissues. Accurate modeling of these propagation dynamics is essential for understanding the systemic consequences of molecular interventions.

With such pretrained embeddings for genes and proteins, we can facilitate a variety of higher-level modeling applications. For instance, these embeddings can provide positional biases when modeling gene expression within cells, enhancing the representation of spatial and regulatory contexts. Additionally, they can serve as initial embeddings for simulating the propagation of perturbation effects through molecular networks, thereby improving the predictive accuracy of models that aim to understand cellular responses to interventions. This approach not only enriches the representation of individual molecular entities but also enables the integration of network-level information into downstream predictive and generative models. By capturing the complex interplay among genes and proteins within biological networks, we can advance our ability to model biological processes more holistically and design more effective therapeutic strategies.

In such a model, the initial embedding of each molecule can be provided by molecular-level embedding models for their respective sequences or structures. Then these embeddings can be further transformed by the graph neural network to take into account the network effects. For pretraining such graph neural networks, contrastive learning objectives can be employed. Specifically, contrastive loss functions encourage the model to produce similar embeddings for entities that are proximal in the network, while assigning dissimilar embeddings to entities that are more distant. This approach enhances the model’s ability to reflect the underlying network topology in the learned representations. 

\textbf{Incorporating lower-level embedding into higher-level modeling}. In an AIDO, lower-level representations, such as molecular embedding, can be used in models for a higher level, such as a cell. Current single-cell RNA sequencing (scRNA-seq) models typically handle gene expression data by treating each gene as a discrete symbol, neglecting the rich sequence information inherent to each gene. In transformer-based pretraining approaches, embeddings for each gene are learned exclusively from gene expression levels, which does not leverage the extensive biological information encoded in gene sequences. Additionally, existing models often overlook established biological knowledge regarding gene regulatory networks and protein–protein interactions. In an AIDO, 
we propose two key architectural innovations. 

\begin{figure}
    \centering
    \includegraphics[width=0.9\textwidth]{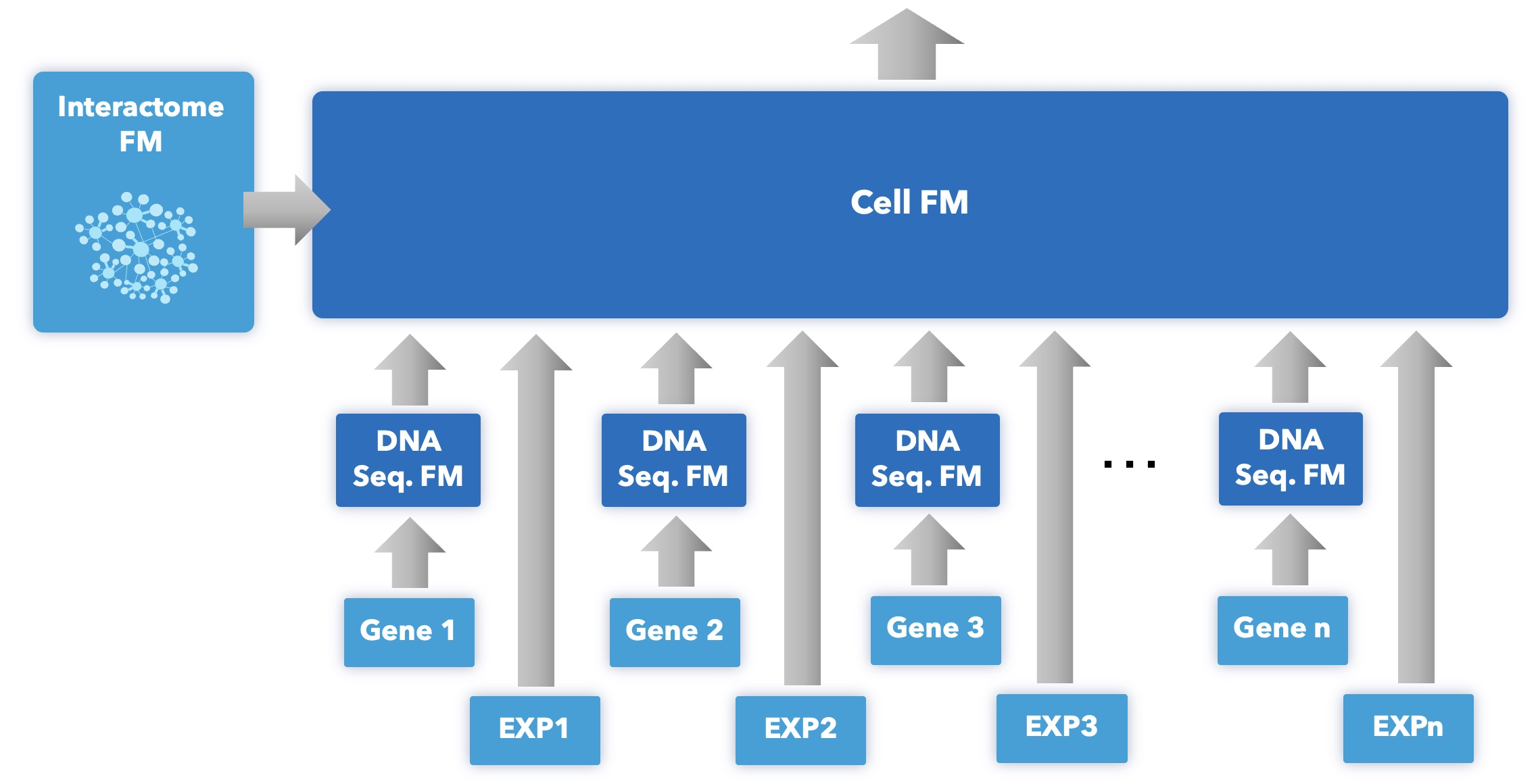}
    \caption{FM for Cell which can leverage embeddings from DNA sequence FM and Interactome FM.}
    \label{fig:fm4cell}
\end{figure}

First, embeddings derived from both the regulatory sequence and the coding sequence of each gene shall be used as the positional embeddings in the model (Figure~\ref{fig:fm4cell}). By assigning two distinct embeddings to each gene—one representing the regulatory sequence and another representing the coding sequence—we can capture different aspects of gene expression regulation. Genes with similar regulatory sequences may be co-regulated by the same transcription factors, influencing their expression patterns, while genes with similar coding sequences might exhibit comparable transcriptional and translational efficiencies. This dual-embedding strategy allows the model to disentangle the regulatory influences from the effects of coding sequences, providing a more nuanced representation of gene expression dynamics.

Second, known gene regulatory networks, protein–protein interaction networks, and gene function networks shall be used to introduce positional embedding biases within the transformer’s attention mechanism. By leveraging these networks, we encode prior knowledge about the relationships and interactions among genes directly into the model. Genes connected as neighbors in these networks can inform the model about potential homophilic (similar genes influencing each other) or heterophilic (dissimilar genes influencing each other) effects on gene expression. We apply a graph neural network to these heterogeneous networks to generate embeddings for each gene, which are then used to bias the attention matrix in the transformer. This integration effectively modifies the transformer embeddings to reflect known biological interactions, enhancing the model’s ability to capture complex regulatory patterns and improving predictive performance. Furthermore, when multi-omics data are available for the same cells—such as protein expression profiles from proteomics or chromatin accessibility data from ATAC-seq—they can be integrated into the model. This integration can be achieved either by extending the single-cell model to incorporate additional data types or by aligning datasets across different modalities. Including these additional layers of information provides a more comprehensive view of cellular states and can significantly improve the model’s accuracy and generalizability.

The idea of using low-level embedding for the high-level model goes beyond gene-to-cell interactions. For example, image data can also be used to incorporate morphological features for even higher-level modeling. It lays the foundation for a comprehensive modeling strategy to understand and predict cellular behaviors, which can have significant implications in developmental biology, disease mechanisms, and therapeutic interventions.

\textbf{Integrating FMs across scales}. Molecular FM, cellular FM, and phenotype FM, each enable AI-driven functions at their respective biological scale. An AIDO  goes further by building an integrative system using these FMs as building blocks to modeling tasks at the organ, organ network, and individual levels for a comprehensive understanding of complex biological systems and diseases. 
Large-scale population cohorts such as the UK Biobank and the Human Phenotype Project (HPP) are making available extensive phenotypic data, including longitudinal and time-series measurements, as well as molecular and cellular measurements, which can be leveraged to build robust models given sufficiently large and standardized cohorts. 

This health data enables us to link molecular and cellular presentations to observable traits and clinical outcomes, effectively bridging the gap between genotype and phenotype. Complex phenotypic measurements—such as continuous glucose monitoring patterns, sleep quality metrics from wearable devices, dietary intake logs, and physical activity levels tracked by accelerometers—can be tokenized by autoencoder architectures and other deep learning techniques specialized for time-series data as we discussed earlier. These models learn latent representations that capture essential features influencing health and disease states. Imaging data collected from individuals, such as retinal scans, magnetic resonance imaging (MRI), computed tomography (CT) scans, and other modalities, provide rich phenotypic information reflecting underlying molecular and cellular processes.

\begin{figure}
    \centering
    \includegraphics[width=0.7\textwidth]{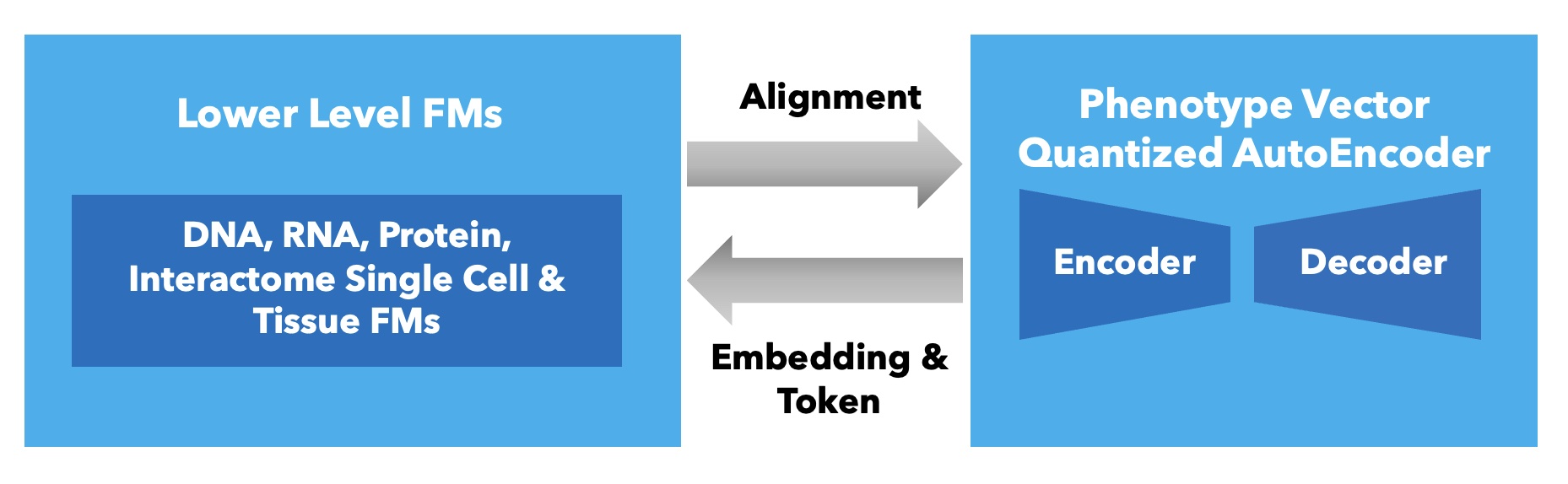}
    \caption{FM for phenotype data leverages autoencoder to model phenotypic information which can inter-operate with lower level FMs for molecule, cell and tissues.}
    \label{fig:phenotype}
\end{figure}

By integrating these phenotypic models with foundation models from lower biological scales—such as molecular-level models of gene expression and protein interactions and cellular-level models of cell types and states—we can create multiscale models that capture the complexity of biological systems (Figure~\ref{fig:phenotype}). This integration can be achieved by aligning the latent representations learned at each level. For example, embeddings from molecular foundation models can be connected with phenotypic embeddings through shared variables like gene expression profiles or genetic variants identified in genome-wide association studies (GWAS). Connecting the lower and higher-level representations also allows us to obtain mechanistic insights into the molecular underpinnings of phenotypic observations.

By connecting phenotypic models with foundation models from molecular and cellular levels, we establish a holistic framework capable of capturing biological complexity across scales. This approach enables the exploration of how molecular and cellular alterations manifest at the phenotypic level, facilitating better predictions of disease progression, treatment responses, and personalized therapeutic interventions. Furthermore, this integrated modeling strategy allows for the inclusion of environmental and lifestyle factors, providing a more comprehensive picture of health and disease. By accounting for the interplay between genetics, biology, behavior, and environment, we can move towards truly personalized medicine that tailors interventions to the unique profile of each individual.

\subsubsection{Piece It All Together: Align and Optimize across Scales}

\begin{figure}
    \centering
    \includegraphics[width=0.98\textwidth]{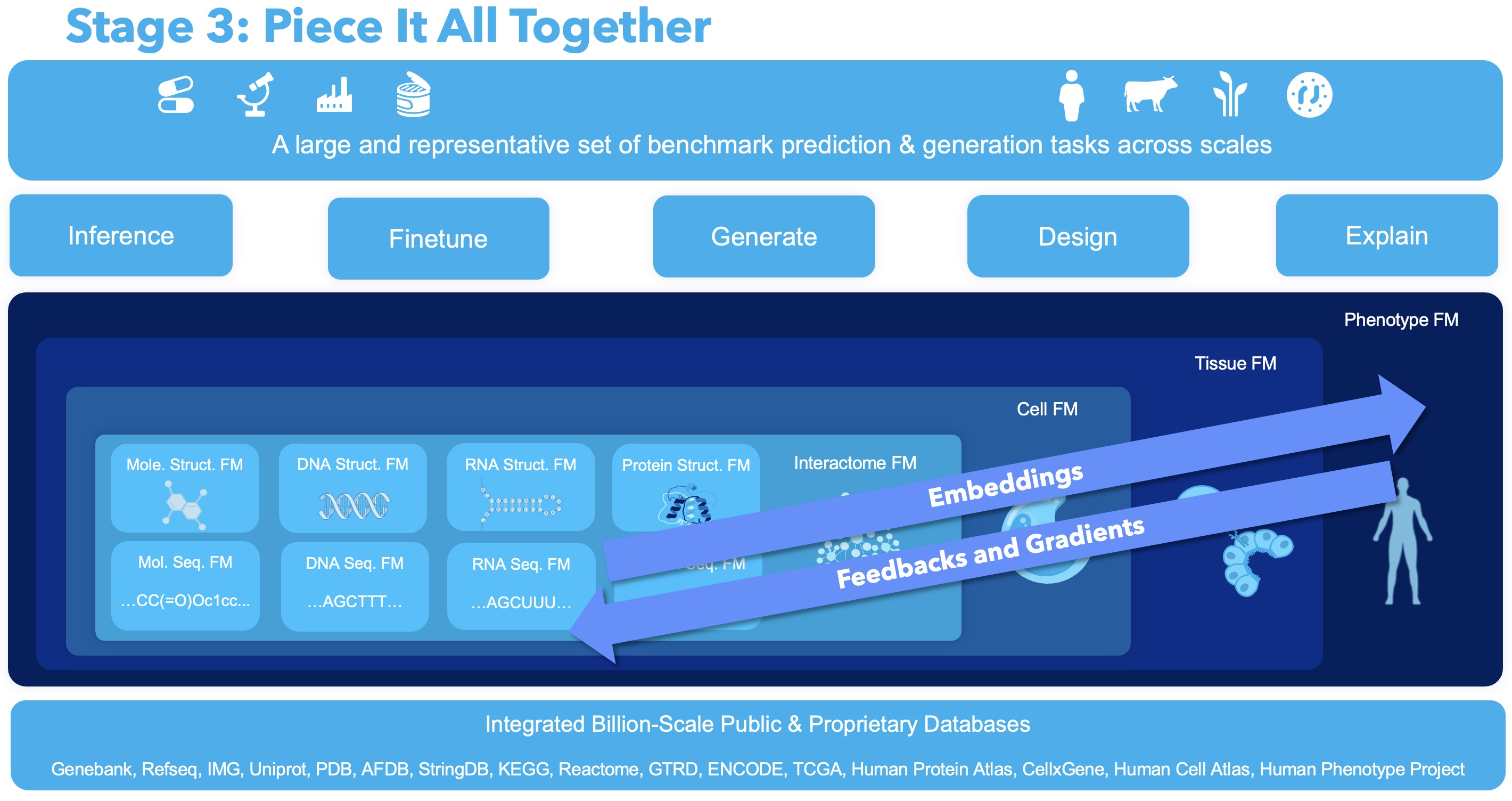}
    \caption{Stage 3 of building an AIDO.}
    \label{fig:mulstiscalefm_stage3}
\end{figure}

Bringing all of the components above together allows the proposed biological foundation models to intricately integrate the central dogma and the multiscale, interconnected nature of biology into architectural designs. 
By \textbf{integrating phenotypic models with foundation models from molecular and cellular levels}, we establish a holistic framework capable of capturing biological complexity across scales (See Figure~\ref{fig:mulstiscalefm_stage3} for a summary). Collectively, these architectural innovations demonstrate how foundation models can integrate multiscale biological information—from molecular sequences, to cellular contexts and evolutionary relationships, and to the individual organism and its temporal behavior—providing a cohesive framework for modeling complex biological systems. This integrative approach holds the promise of advancing our ability to simulate and understand biology across different scales, ultimately contributing to breakthroughs in biomedical research and personalized medicine.

One key feature of the modules in an AIDO system is that they are connected to build new models or modules, and high-level models are constructed hierarchically and are nested with lower-level modules. Such nested construction and vectorized and differentiable connections do not only allow representations of information to flow from the bottom molecular level to the higher phenotype level, but also allow feedback information to flow back in the architecture to use higher level information to further adjust, optimize or align lower level embeddings and models. Here we have the opportunity to leverage the supervision signals from a multitude of predictive and generative tasks across different scales to jointly adjust or optimize all modules in an AIDO together, to make them aligned and become a truly unified whole. 

Since the modules from each level and for each biological data modality have already been pretrained with their own objectives, and adjusted by the important downstream tasks corresponding to their biological levels, we will need the overall unified system to respect and keep consistency with these objectives. Thus, when we jointly adjust and align the system modules, we will use a weighted combination of all these task objectives and jointly optimize these objectives, which allows the modules to be adapted such that they are more aligned with each other and the overall performance of the entire AIDO system can be improved. 

\section{How to Use an AI-Driven Digital Organism}

To achieve the capabilities of prediction, simulation, experimentation, programming, evolution, and optimization with multiscale foundation models, it is essential to develop a suite of techniques and software to facilitate model building based on the system of multiscale foundation models. 

These techniques will help us to fully harness the potential of the system of multiscale foundation models to address critical challenges in biological discovery and engineering. Below we will explain these techniques in more detail, and provide concrete examples on how they can be used for biological applications in different scales. 

\subsection{Techniques for Adapting an AIDO for Downstream Tasks}
\label{sec:techniques_adapt}

The set of pretrained models in an AIDO is optimized for general objectives and with a broader set of pretraining data. Although these pretrained models and the set of embeddings from these models may be generally good representations, they are specialized to a particular family of data or directly built for the biological problems at hand. Therefore, adaptions are often needed to turn these pretrained generalist models to high-performing models for the data and tasks at hand. To achieve efficient adaption or specialization in downstream tasks, it is also crucial that only a small amount of computation or data is used in the process. Another common characteristic of these techniques is that only a smaller number of additional parameters are added to the overall models, and the pretrained models are changed only slightly. 

\textbf{Regression and Classification} 
For straightforward regression and classification tasks, such as mRNA stability prediction, protein expression prediction and cell type predictions, predictive models can be constructed by adding a few layers of neural networks on top of the embeddings generated by the pretrained models. With a modest set of labeled data points, the weights of these additional layers can be trained, and the weights of the pretrained model can also be adjusted using memory-efficient methods such as Low-Rank Adaptation (LoRA)~\cite{hu2021lora}. This approach enables the models to adapt to specific tasks without the need for extensive computational resources. For example, the embeddings of a foundation model constructed on time-series continuous glucose monitoring data can be fine-tuned by linear regression to predict clinical measures such as blood pressure and various blood biomarkers.

\textbf{Continual Pretraining} 
Another effective strategy for adapting pretrained foundation models is continual pretraining, which involves further training the models on data distributions that are more closely aligned with the downstream tasks~\cite{gupta2023continual}. While the initial pretraining data may cover a broad distribution, downstream tasks often require the model to focus on a local distribution specific to a particular family of species or cell types. Continual pretraining leverages unlabeled data relevant to the downstream tasks, allowing the model to refine its representations without overfitting, and requires relatively little computational effort. For instance, a DNA sequence foundation model may be pretrained over sequences from all species including prokaryotic and eukaryotic species; when we want to apply it to a primate genome prediction problem, continual pretraining with primate genomes will help the foundation model to better specialize to represent such data. Another example is a foundation model for the human cell which can be continuously pretrained to better present various types of immune cells. 

\textbf{Augmentation with Auxiliary Models}
For more complex downstream tasks, such as protein and RNA structure prediction, more sophisticated models are necessary. These tasks involve intricate structural relationships that are best captured using deep architectures like convolutional neural networks, Evoformers~\cite{jumper2021highly}, and specialized structural modules. These architectures further transform the embeddings from the pretrained models to accurately predict 3D structures, which are essential for understanding biological functions and interactions.

\textbf{Fusion Models} 
For some complex tasks, such as protein isoform expression prediction and cell perturbation response prediction, multiple data inputs and their representations need to be aggregated to build a model. For instance, protein isoform expressions are controlled by the regulatory region in genome sequences, the intron and exon sequences for alternative splicing, and also the cellular state and the codon sequence of the corresponding protein isoform. Therefore cell or tissue-specific protein isoform expression models need to take into account both DNA, RNA and protein information. In these cases, various ways of fusing pretrained models, such as early and late fusion techniques in machine learning~\cite{baltruvsaitis2018multimodal}, and specific deep models to do that, such as cross-attention networks or gated embedding aggregations, need to be defined to fuse multiple pretrained models to develop downstream task models. 

\textbf{Conditional Generation Models}
Conditional generation models can also be developed based on the pretrained language models. For example, one can perform conditional generation of entire protein sequences based on specific sequence motifs. In such cases, a denoising discrete diffusion model can be trained, using the pretrained models as initialization for the denoising networks~\cite{alamdari2023protein}. Another example of conditional generation is molecular inverse folding, where the objective is to predict a protein or RNA codon sequence given its basic backbone structure. In this scenario, a structure encoder is employed to represent the input structures, and the embeddings produced are utilized in a cross-attention mechanism to influence the output of a pretrained sequence model. This approach enables the generation of sequences that are compatible with specific structural constraints, facilitating the design of molecules with desired properties.
These methods allow for the generation of novel sequences that retain desired features or functionalities, which is particularly valuable in drug discovery and enzyme engineering. 

In summary, by employing these adaptation techniques and integrating architectural components across different biological scales, we can harness the power of foundation models to address a wide range of biological design problems. This approach enables us to predict, simulate, experiment with, program, evolve, and optimize biological systems, ultimately advancing our understanding and manipulation of complex biological phenomena. In the following, we will outline three concrete applications where multiscale foundation models can be used to address interesting questions arising from molecular level, cellular level and phenotype level. 

\subsection{Molecular Engineering}

We begin with an example of how foundation models can be utilized for designing molecules. For instance, to design molecules such as antibodies or enzymes, two types of task-specific models based on foundation models are essential. The first type involves \textbf{conditional generative models}, where candidate protein sequences are generated given a specific design goal—for example, binding to a particular target epitope or achieving certain stability metrics. The second type comprises \textbf{protein property prediction models}, including for instance protein structure prediction, antigen-antibody binding affinity prediction, and protein stability prediction. Both types of models can be developed on top of pretrained foundation models.

\begin{figure}
    \centering
    \includegraphics[width=0.9\textwidth]{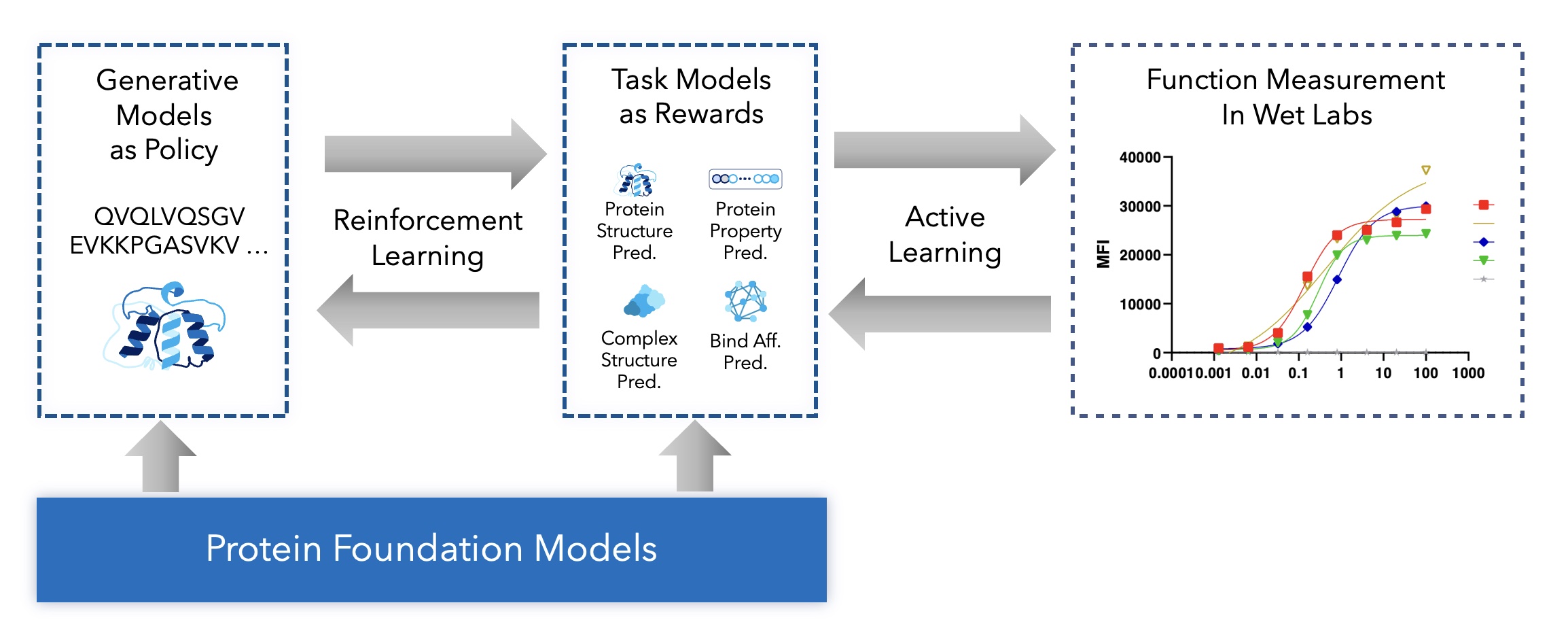}
    \caption{Using foundation models for protein sequences and structures in an AIDO for generative design and engineering of molecules.}
    \label{fig:proteindesign}
\end{figure}

Conditional structure generation can be achieved by adapting a structure decoder, while protein sequence generation can be performed by guiding a protein language model conditioned on the generated structures. These generative models act as policies to produce initial designs and must balance both diversity and quality in the generated sequences.

Once protein structures and sequences are generated, it is crucial to evaluate these designs to identify the most promising candidates. This evaluation involves assessing whether the predicted structure of a generated protein sequence is consistent with the structure used to condition the sequence generation model. Additional scoring functions—such as predictions of protein stability and binding affinity to target molecules—are necessary to further discriminate among the generated designs. All these scoring functions, including complex protein structure prediction models, can be built upon pretrained foundation models for proteins. For instance, structure prediction models can be developed by aligning protein language models with structure decoders through full fine-tuning, and protein stability and binding affinity models are regression models and can be constructed by fine-tuning protein language models using labeled data.

With these protein generative models and scoring models in place, the generative models can be used as policies to generate new protein structures and sequences. The property scoring models serve as discriminators or reward functions to filter the generated designs, guiding the generation process towards proteins with improved metrics. The designed proteins can be selected via active learning, and then be synthesized and tested experimentally in the wet lab. Data obtained from these experiments can be fed back into the generative and discriminative models for further fine-tuning, facilitating subsequent rounds of in silico design and optimization~\cite{chen2024xtrimopglm} (See Figure~\ref{fig:proteindesign} for an illustration). 

\subsection{Cellular Engineering}

Foundation models can also be utilized for cell engineering scenarios, such as immune cell activation or suppression, stem cell differentiation and aging reversal cell engineering.  In order to engineer cell types to some desired status including transcriptomic, morphology or function level, we often need to apply external perturbations to the cells. However, the type of perturbations and the amount of perturbation to a cell can vary. For instance, we can apply small molecule perturbation, antibody perturbation, genome editing or RNA interference to a cell. Furthermore, we can also apply a combination of perturbations to the cell, and perturbation dosage is also important. Thus, we need to develop models for predicting the effect of perturbations on cells and use the model to guide cell engineering. 

\begin{figure}
    \centering
    \includegraphics[width=0.9\textwidth]{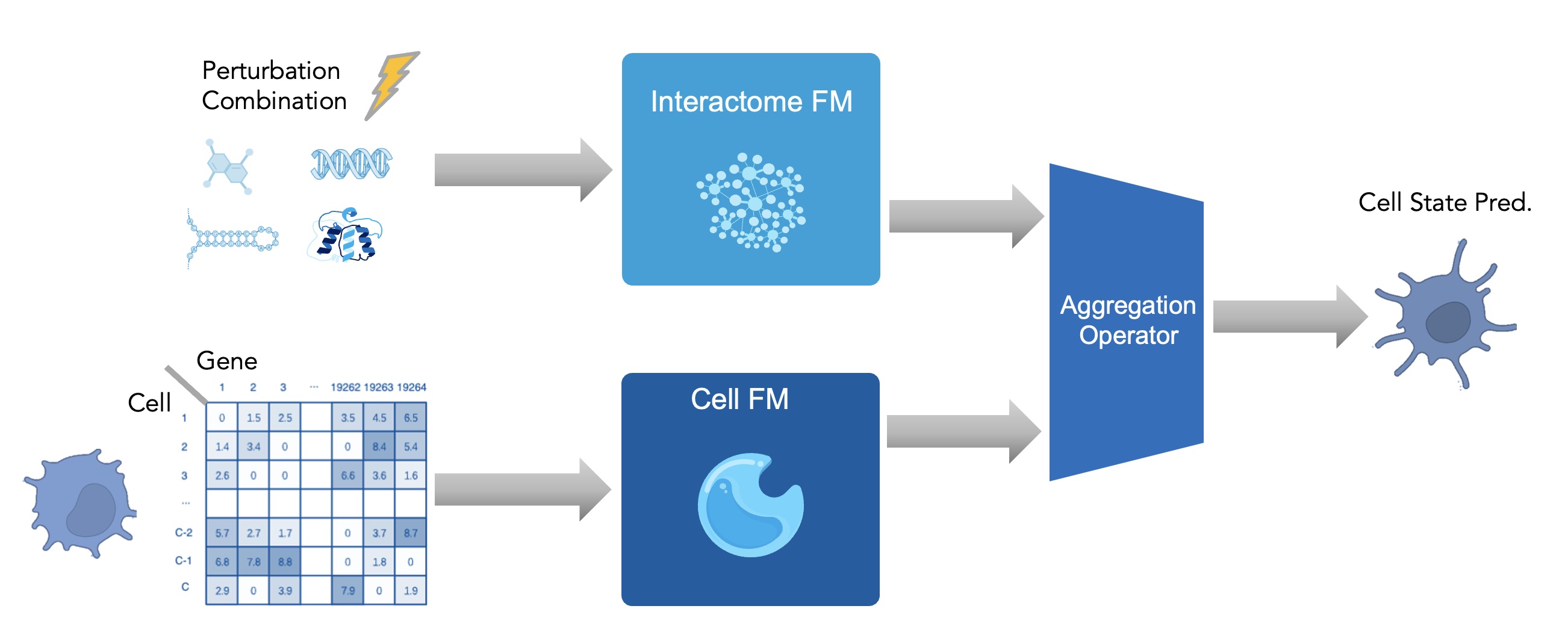}
    \caption{Using foundation models for interactomes and cells in an AIDO to improve cell perturbation prediction.}
    \label{fig:cellperturbation}
\end{figure}

The effects of the small or large molecule or genetic perturbation will propagate in the complex gene regulatory network and various pathways. Therefore, it is important to model the network effects of these perturbations. We can model gene regulatory networks and signaling pathways as a signed network with heterogeneous types of nodes and edges and model the perturbation on these nodes as embeddings produced by graph neural networks. Graph neural networks for heterogeneous graphs allow us to take the node type, edge type (including positive or negative regulation) and graph topology into account and encode such information into vector space representations to connect with downstream prediction models. Such embeddings can capture similarity such that two nearby nodes that are connected with a positive path in a network have similar perturbation effects and their embeddings are thus similar; Conversely, if two nodes are connected with negative regulatory paths then their embedding can be very different even if they are close in the network; perturbation to a genetic node can have different effects from that to the corresponding protein, since deletion of a gene may affect multiple functions carried out by a protein, but the perturbation to a particular protein isoform may just block one of its functional pathways. The representation of the perturbation can be obtained by adapting the foundation models for interactomes.    

Once we have obtained these representations of the perturbations, we can integrate such vectors with the pretrained embeddings for a single cell. The pretrained embedding for a single cell before perturbation essentially provides a representation of the status of the cell before the perturbation. When the perturbation vectors are aggregated with such cellular vector representation, the modified vectors are used to represent the embeddings after the perturbations. With additional multilayer perception models, we can then connect these vectors for gene expression predictions or entire cellular level predictions to life/death/activation level predictions of a cell (Figure~\ref{fig:cellperturbation}). 

With such a prediction model for cellular responses to external perturbations, we essentially obtain an in silico ``simulator'' for cells to answer \textit{what if}  type of questions, allowing us to perform more perturbation experiments in silico.  
In our previous work~\cite{hao2024large}, we show that such an architecture for perturbation prediction can lead to substantial improvement in prediction accuracy even for cases of perturbation combinations that contain unseen genetic perturbations.  Furthermore, such a model will allow us to find potential targets or target combinations to treat cancerous cells, to activate immune cells, to steer cell fate such as reversing the aging process of a cell, or to differentiate a stem cell into a neuron. 

\subsection{Phenotypic  Engineering}

Large-scale population cohorts are generating increasing amounts of clinical and phenotypic data coupled with molecular multi-omics data. Here we give an example of how foundation models at multiple levels can be integrated with such human cohort data, thereby conferring higher-level phenotype capabilities to the AIDO. The Human Phenotype Project (HPP) that we established offers a rich resource characterized by its depth and diversity of phenotypic and health-related information for each participant. This includes multi-omic data, physiological assessments, imaging, continuous health monitoring via wearables, electronic health records, lifestyle indicators, and biobanked biological samples for future analyses. Such comprehensive datasets provide an opportunity to link foundation models across different scales to address diagnostic, health monitoring, and aging-related applications.

\begin{figure}
    \centering
    \includegraphics[width=0.8\textwidth]{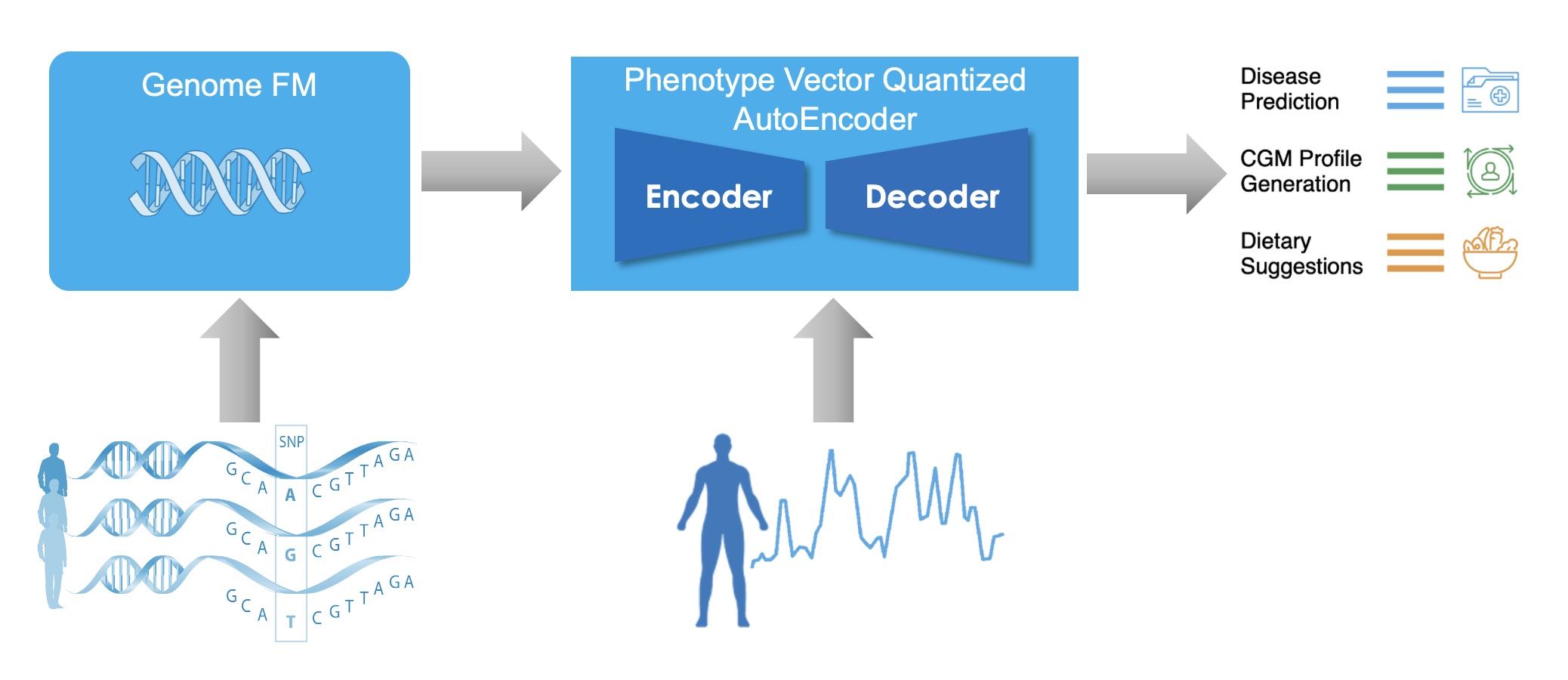}
    \caption{Linking DNA foundation models in an AIDO to continuous glucose response foundation models in an AIDO to perform association study.}
    \label{fig:cgmgenome}
\end{figure}

HPP houses one of the largest collections of continuous glucose monitoring (CGM) data, providing rich temporal information on glycemic patterns. We leveraged this data to develop GluFormer~\cite{lutsker2024glucose}, a generative foundation model based on a transformer architecture, trained on over 10 million CGM measurements from 10,812 non-diabetic individuals. GluFormer was trained using next-token prediction in a generative, autoregressive manner and generalized effectively to 15 external datasets from five geographical regions, six different CGM devices, and various metabolic conditions, including normoglycemic, prediabetic, and diabetic populations. Demonstrating the power of foundation models, GluFormer produced embeddings that outperform traditional CGM analysis tools and achieved high Pearson correlations in predicting clinical parameters such as HbA1c, liver-related markers, blood lipids, and sleep-related indices. Notably, GluFormer can predict the onset of future health outcomes up to four years in advance. By integrating dietary data, we enhanced GluFormer to accurately generate CGM data based solely on dietary intake, simulate outcomes of dietary interventions, and predict individual responses to specific foods. To explain the observed glucose response patterns, we are now building models that leverage pretrained genomic and cellular transcriptomic foundation models. In particular, the embeddings or latent spaces from these lower-level pretrained models can be aligned to the latent spaces of the phenotype model Gluformer using subjects who have simultaneous measurements of phenotype information and genetic and/or PBMC information. In the case of aligning genomic information and phenotype information, multiple single nucleotide polymorphisms (SNPs) for a gene—including coding and regulatory regions—can be embedded using a pretrained DNA model, and then linked to CGM outputs, potentially improving statistical power by avoiding multiple hypothesis comparisons that treat each SNP as an independent variable (Figure~\ref{fig:cgmgenome}).

Similarly, we also developed a foundation model for electrocardiograms (ECG) that integrates genetics data, uncovering novel insights into cardiovascular health. We used contrastive learning to train a self-supervised model on ECG data from 4,782 HPP subjects to derive temporal shifts in cardiovascular state, termed Delta ECG~\cite{levine2024genetic}. By carrying out a genome-wide association study (GWAS), on these ECG-derived representations, we identified genetic signals associated with temporal changes in cardiovascular health, including five genome-wide significant associations. Additionally, our model embeddings predicted the expression of genes from Peripheral Blood Mononuclear Cells (PBMCs), revealing significant enrichment in pathways related to the electron transport chain and immune responses. This integration of genetics and cardiovascular dynamics showcases the power of combining phenotypic and molecular data, in this case enabling the identification of previously unexplored biological mechanisms underlying cardiovascular disease.

\section{Scalable Computing Infrastructure and Software}

\begin{figure}
    \centering
    \includegraphics[width=0.9\textwidth]{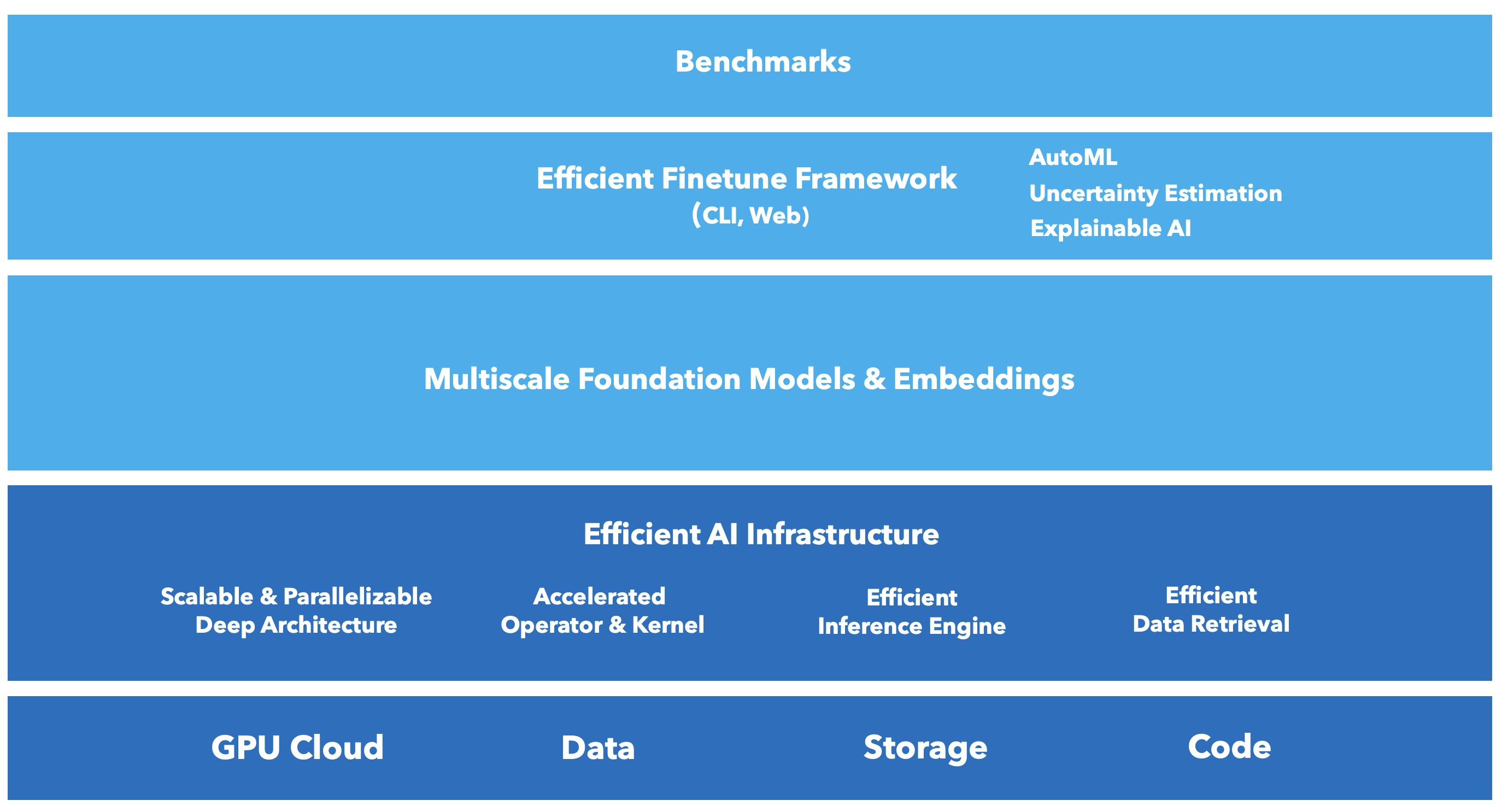}
    \caption{Scalable computing infrastructure and software for an AIDO.}
    \label{fig:system}
\end{figure}

The number of parameters, the amount of data, and the complex interconnectedness in the multiscale foundation models underlying an AIDO pose significant challenges for existing computing infrastructures and systems. Furthermore, in order for computational biologists, medical researchers, experimental biologists, and drug designers to use the AIDO, necessary software packages also need to be developed to facilitate its usage (Figure~\ref{fig:system}). Below we offer our view on the technical routes to address these challenges.  

\subsection{High-Performance Computing System for Biology}

Since the number of parameters and the amount of data needed to pretrain the FMs in an AIDO are in the scale of billions, and will continue to grow, industrial-scale computing hardware such as large GPU-clusters are needed. Furthermore, the system and software infrastructures also need to incorporate various parallel and high-performance computing techniques to speed up the training process~\cite{shoeybi2019megatron,rasley2020deepspeed,touvron2023llama}. For instance, if there are many data points in a batch, they can be processed in parallel in different GPUs using data parallelism. If the model parameters or embedding vectors are very large, these tensors will need to be distributed across multiple GPUs with tensor parallelism. If input sequences are too long, we also need to use sequence parallelism where each GPU only processes part of the sequence in parallel. If there are too many layers for a deep architecture, then different layers can be partitioned to different machines such that these layers can be computed in a pipeline fashion to reduce the overall runtime. During training, model checkpoints will need to be saved periodically, such that when there are machine break-downs or when the training diverges, we can roll the model back to the latest normal checkpoint, skip the problematic data, and continue the training. 

Furthermore, jointly aligning all different pretrained models in stage 3 of an AIDO after connecting them together will also be very challenging. This will require the development of new and efficient systems that make the joint adaptation of these interconnected foundation models possible. Systems that can distribute the forward inference and backward feedback according to the connectivity of the hierarchical and nested system of models yet which can manage the communication efficiently will be crucial for building a truly unified AIDO system. 

Besides parallel and distributed computing, hardware-level programming techniques that can leverage hardware properties will also be needed. For instance, fusing some commonly used sets of operators in the models and developing hardware-level kernel codes for such fused operators will also help accelerate the computation. Furthermore, in the implementation of some kernel codes, different tiers of caching with different speeds and sizes can also be taken into account to gain further speed up, such as the FlashAttention~\cite{dao2022fast}. 
Finally, for speeding up model training and inference, mixed precision computing~\cite{micikevicius2017mixed}, model compression~\cite{hinton2015distilling} and compiled inference codes can help develop efficient inference engines.   

\subsection{Standardized Software Stack and Development Tools}

A major bottleneck in the field of foundation models for biology is the very chaotic and fragmented production and adoption of software tools for data processing, pretraining, fine-tuning, and evaluation. A wide range of public and homemade code-bases originated from diverse sources like Hugging Face\footnote{\url{https://huggingface.co/}}, Megatron\footnote{\url{https://github.com/NVIDIA/Megatron-LM}}, DeepSpeed\footnote{\url{https://github.com/microsoft/DeepSpeed}}, can be found in the publications, with different data format, library dependencies, versioning, hardware specifications, etc. being present.  

In order for practitioners to easily use, compare, and experiment with the pretrained models from an AIDO and elsewhere, an efficient and versatile software package for adapting the models in an AIDO at large in a standardized and unified way is needed. We are developing such a package which will include certain model checkpoints and various model adaptation techniques mentioned in \S~\ref{sec:techniques_adapt} using command lines, APIs, or web user interfaces. The package will be released under the repository
\begin{center}
    \url{https://github.com/genbio-ai/AIDO}    
\end{center}
Initially, the repository will also include the AIDO building blocks, such as state-of-the-art component FMs we have trained from scratch, including: 
AIDO.DNA~\cite{ellington2024dna}, 
AIDO.RNA~\cite{zou2024rna}, 
AIDO.Protein~\cite{sun2024mixture}, 
AIDO.StructureTokenizer~\cite{zhang2024balance} and 
AIDO.Cell~\cite{ho2024scaling}; and some advanced models leveraging AIDO system's ability to connect multiple building blocks such as 
protein inverse folding (AIDO.ProteinIF~\cite{sun2024mixture}), 
RNA inverse folding (AIDO.RNAIF~\cite{zou2024rna}) and 
retrieval augmented protein language model and structure prediction model (AIDO.RAGPLM and AIDO.RAGFold~\cite{lee2024ragfold}). We will continue to update the repository for AIDO by including more pretrained building blocks, upgrading existing modules, assembling advanced models using multiple building blocks, and then eventually the integrated and unified AIDO system. 
Furthermore, AutoML techniques will also be incorporated to support automating the process of hyperparameter search for downstream tasks. Additionally, functionalities such as uncertainty quantification and model explanations, will be incorporated to make the adapted downstream models more trustable and easy to understand. 

Representative benchmark tasks at different scales of the biological system are of paramount importance to drive continuous improvement of the AIDO models, individually and as a whole. 
Such a systematic array of benchmarks from molecular level to cellular level to individual level needs to be compiled from public databases as well as from potential new experimental sources. For instance, for protein level tasks, there are ProteinGym benchmarks\footnote{\url{https://proteingym.org/benchmarks}}; for cellular level tasks, there are scPerturb\footnote{\url{https://projects.sanderlab.org/scperturb/}} and GDSC\footnote{\url{https://www.cancerrxgene.org/}}; for phenotype level tasks, there are UK biobank\footnote{\url{https://www.ukbiobank.ac.uk/}} and HPP\footnote{\url{https://humanphenotypeproject.org/home}}. These datasets need to be standardized and integrated into a pipeline automated for large-scale evaluation. So far, we have evaluated various components of an AIDO over a collective set of more than 300 tasks, and showed that the models in an AIDO are achieving overall SOTA performance in various aspects across various scales
~\cite{
ellington2024dna, 
zou2024rna, 
sun2024mixture, 
zhang2024balance,
ho2024scaling,
lee2024ragfold}.  
Besides these existing benchmark datasets, continuous efforts are needed to further optimize the benchmarks and evaluation schemes for sustainable model improvement, which we expect our software tools will endeavor to support. 

\subsection{Open-Source Provision and Ecosystem}

Realizing the AIDO vision requires long-term, continuous, and sustainable development and community efforts.
In fostering such an effort,  
we shall make the weights of certain matured version of the models and the adaptation software packages publicly available for reproducibility, community building, and standardization. The goal is to connect life science, medicine, pharmacy, and public health through a shared technical paradigm, with a wider community involvement including academia, industries and governments, aligned on purpose-driven and coordinated massive data generation and collection efforts, and close-loop collaboration on the full cycle of data, model, hypothesis, outcome, back to more-data generation.

Through continuous versioning and upgrade of the base models, APIs, task-suites, data banks, and bio-entity representation repositories, fueled by the ever-increasing willingness of data sharing and federated data mining, and the never-fading demand for synthetic biology and personalized medicine, we believe a new community of users and developers of an AIDO can emerge and grow from this open-source effort as we have seen in the LLM field, to together pursue the common vision of mirroring life in the physical world on a computer with AI. 

\section{Benefits and Comparisons to Alternative Approaches}

It can be seen from the above that by building an AIDO, a system of integrated multiscale foundation models, many biological problems can be 1) better addressed in a unified framework rather than with each specialized model as seen in the literature already; 2) all the models can be built in a standardized, one-stop software toolkit; 3) modeling and system innovations such as newer architecture, newer modules, efficient training and inference techniques occurring in one spot of the system can potentially be used to improve other modules and hence the entire effectiveness and efficiency of the system. Such a system has obvious advantages as compared to other alternative methods. 

Under the influence of modern physics and chemistry, biological science, traditionally a heavily empirical and experimental science, has made great strides in our understanding of ``What Is Life'' with rationalism and reductionism based methodologies, much like how astronomy understands and describes the celestial system using physical laws. However, because of the extreme complexity of the biological systems due to the myriad of elements interacting at atomic, molecular, cellular, tissue, organ, organismal, and even social levels, and the colossal volume of biological data modern technologies are able to collect, first principle methods based on physical laws, symbolism, and computationalism thereupon scale poorly to offer predictive and actionable understanding and intervention capability in many biological problems, such as disease diagnosis, drug design, etc. 

\textbf{Molecular dynamics simulations} can capture the physical and chemical properties of small molecules with high precision. However, they fall short when it comes to simulating and predicting phenomena at larger scales. The computational resources required to model the vast number of entities involved at the cellular level and beyond become prohibitively large, rendering molecular dynamics impractical for simulating larger-scale biological systems.

\textbf{Rule-based systems}, which rely on precise knowledge and logical reasoning to verify existing information and derive new hypotheses, are also limited. The current state of biological knowledge is significantly incomplete, restricting the versatility and generalizability of such systems. Without comprehensive rules that encapsulate the complexities of biological processes, these systems cannot adequately model the full spectrum of biological phenomena.

\textbf{Task-specific models}, such as those used in genome-wide association studies (GWAS) or regulatory network analyses, are developed to address a single well-defined problem. While these models can be effective for their intended purposes, they are typically constructed using small, highly specialized datasets. This specificity limits their transferability to other scenarios and confines their applicability to narrow domains. Moreover, each model usually addresses a problem at a specific scale, further limiting its utility across the broader spectrum of biological complexity.

In contrast, an AIDO approach provides what can be called an ``actionable empirical understanding'' of the subjects, builds on a wide range of machine learning computations such as self-supervised learning, on large amount of data, in the form of predictive and generative models that connects data input to task output that nature would have produced, in a robust, repeatable, and verifiable, but not necessarily explainable way, which contrasts what rationalism and reductionism based methods would expect, such as laws and theories. Furthermore, the system of foundation models in an AIDO can be rapidly adapted to a wide range of downstream tasks and can be combined and cascaded as modules to build more complex models. This adaptability allows an AIDO to overcome many of the limitations associated with other methods. Furthermore, leveraging our knowledge of how biological entities from different scales are interconnected, and how the data are generated and related, inductive biases and rich prior information can also be incorporated in the system of pretrained models to achieve a good unification and balance on knowledge-driven approach and data-driven approach. By harnessing the wealth of unsupervised data available and the biological domain knowledge, the system of foundation models in an  AIDO provides a versatile and scalable approach to modeling biological systems across multiple scales. The arrival of an AIDO opens the door to a new wave of connectionist revolution in the empirical study of biological subjects, allowing many prediction problems in biology with small amounts of task labels to improve substantially, redefining ``computing'' in biology study. 

\section{Other Considerations and Outlook}

Besides the above technical aspects on how to build the AIDO or multiscale foundation models, there are many other aspects to consider which we outline below, but leave to a future work for the deep-dive. 

\subsection{Explainability}

Can we extract "causal" logic within the FMs to mechanistically explain the predictive or generative outcome? We can leverage much previous research on explainable methods for deep learning models to address this consideration. For instance, Shapley value and gradient of the output with respect to the inputs can be used to explain which part of the inputs the models are sensitive for decision making~\cite{lundberg2017unified}. 
For fine-tuning and model adaptation, we can apply explainable-by-design architectures such as contextualized models \cite{lengerich_contextualized_2023}, which reveal structured interactions between features and across modalities.

\subsection{Trustworthy}

Are the results reliable and believable? We can work on three important dimensions to build trust: 1) clearly define the production process and usage regions of an AIDO: rely on public large data for pretrain, and leave personalization and specialization based on private data to the adaptation stage with tight data protection; 2) open-source the pretrained models for community to test and reproduce; 3) standardize the data formatting, open source the models, develop software API for easier use and evaluation strategy to keep track of model development trajectories.

\subsection{Safety}

How can we ensure that our models are not abused or maliciously used, such as generating harmful designs, and making controversial predictions? How can we prevent it from being weaponized? Since the design and experimentation are conducted {\it in silico}, it will not directly affect the physical world. Thus regulation can be placed upon physical and wet lab experiments to prevent malicious design and weaponizeable designs from being actually synthesized in the physical world. This is akin to freedom of speech, but laws and regulations govern the physical world actions of a person. 

\subsection{Outlook}

What we are envisioning is a new computational paradigm for addressing life science and biology problems in a holistic fashion. It is an all-purpose tool/paradigm for many use cases arising from different scales or levels of biology. We envision a continuous effort to build up an AIDO or a system of multiscale foundation models for biology. This will involve comprehensive efforts in terms of new deep learning models, comprehensive data generation and integration, better evaluations, 
unified software development for better accessibility, which we envision will be an exciting journey to come. 


\bibliographystyle{unsrt}
\bibliography{ref}

\end{document}